\documentclass{article}
\usepackage[utf8]{inputenc}
\usepackage[T1]{fontenc}
\usepackage[table,dvipsnames]{xcolor}
\usepackage[main,final]{neurips_2026}
\makeatletter
\renewcommand{\@notice}{}
\makeatother

\usepackage[most]{tcolorbox}
\usepackage{fvextra}
\usepackage{microtype}
\usepackage{graphicx}
\usepackage{subcaption}
\usepackage{booktabs} 
\usepackage{makecell}
\definecolor{tablegreen}{HTML}{32a852} 
\definecolor{tablered}{HTML}{cf222e} 
\IfFileExists{bbm.sty}{\usepackage{bbm}}{\newcommand{\mathbbm}[1]{\mathbf{##1}}}
\usepackage{multirow}
\usepackage{multicol}
\usepackage{hyperref}

\usepackage{amsmath}
\usepackage{amssymb}
\usepackage{mathtools}
\usepackage{amsthm}
\usepackage{soul}
\usepackage{fvextra}

\providecommand{\hide}[1]{}

\definecolor{mygreen}{RGB}{202,216,198}
\definecolor{myblue}{RGB}{198,208,219}

\DeclareUnicodeCharacter{2016}{\ensuremath{\Vert}} %
\DeclareUnicodeCharacter{2264}{\ensuremath{\leq}}  %
\DeclareUnicodeCharacter{2265}{\ensuremath{\geq}}  %
\DeclareUnicodeCharacter{2026}{\dots}              %
\DeclareUnicodeCharacter{00B2}{\textsuperscript{2}}%
\DeclareUnicodeCharacter{03BB}{\ensuremath{\lambda}} %
\DeclareUnicodeCharacter{03C3}{\ensuremath{\sigma}}  %
\DeclareUnicodeCharacter{221A}{\ensuremath{\sqrt}}

\usepackage[utf8]{inputenc}
\usepackage{textcomp} %

\DeclareUnicodeCharacter{03B1}{\ensuremath{\alpha}}   %
\DeclareUnicodeCharacter{03B2}{\ensuremath{\beta}}    %
\DeclareUnicodeCharacter{03B3}{\ensuremath{\gamma}}   %
\DeclareUnicodeCharacter{03B4}{\ensuremath{\delta}}   %
\DeclareUnicodeCharacter{03B5}{\ensuremath{\epsilon}} %
\DeclareUnicodeCharacter{03B6}{\ensuremath{\zeta}}    %
\DeclareUnicodeCharacter{03B7}{\ensuremath{\eta}}     %
\DeclareUnicodeCharacter{03B8}{\ensuremath{\theta}}   %
\DeclareUnicodeCharacter{03B9}{\ensuremath{\iota}}    %
\DeclareUnicodeCharacter{03BA}{\ensuremath{\kappa}}   %
\DeclareUnicodeCharacter{03BB}{\ensuremath{\lambda}}  %
\DeclareUnicodeCharacter{03BC}{\ensuremath{\mu}}      %
\DeclareUnicodeCharacter{03BD}{\ensuremath{\nu}}      %
\DeclareUnicodeCharacter{03BE}{\ensuremath{\xi}}      %
\DeclareUnicodeCharacter{03BF}{o}                     %
\DeclareUnicodeCharacter{03C0}{\ensuremath{\pi}}      %
\DeclareUnicodeCharacter{03C1}{\ensuremath{\rho}}     %
\DeclareUnicodeCharacter{03C2}{\ensuremath{\varsigma}}%
\DeclareUnicodeCharacter{03C3}{\ensuremath{\sigma}}   %
\DeclareUnicodeCharacter{03C4}{\ensuremath{\tau}}     %
\DeclareUnicodeCharacter{03C5}{\ensuremath{\upsilon}} %
\DeclareUnicodeCharacter{03C6}{\ensuremath{\phi}}     %
\DeclareUnicodeCharacter{03C7}{\ensuremath{\chi}}     %
\DeclareUnicodeCharacter{03C8}{\ensuremath{\psi}}     %
\DeclareUnicodeCharacter{03C9}{\ensuremath{\omega}}   %

\DeclareUnicodeCharacter{0391}{A}                     %
\DeclareUnicodeCharacter{0392}{B}                     %
\DeclareUnicodeCharacter{0393}{\ensuremath{\Gamma}}   %
\DeclareUnicodeCharacter{0394}{\ensuremath{\Delta}}   %
\DeclareUnicodeCharacter{0395}{E}                     %
\DeclareUnicodeCharacter{0396}{Z}                     %
\DeclareUnicodeCharacter{0397}{H}                     %
\DeclareUnicodeCharacter{0398}{\ensuremath{\Theta}}   %
\DeclareUnicodeCharacter{0399}{I}                     %
\DeclareUnicodeCharacter{039A}{K}                     %
\DeclareUnicodeCharacter{039B}{\ensuremath{\Lambda}}  %
\DeclareUnicodeCharacter{039C}{M}                     %
\DeclareUnicodeCharacter{039D}{N}                     %
\DeclareUnicodeCharacter{039E}{\ensuremath{\Xi}}      %
\DeclareUnicodeCharacter{039F}{O}                     %
\DeclareUnicodeCharacter{03A0}{\ensuremath{\Pi}}      %
\DeclareUnicodeCharacter{03A1}{P}                     %
\DeclareUnicodeCharacter{03A3}{\ensuremath{\Sigma}}   %
\DeclareUnicodeCharacter{03A4}{T}                     %
\DeclareUnicodeCharacter{03A5}{\ensuremath{\Upsilon}} %
\DeclareUnicodeCharacter{03A6}{\ensuremath{\Phi}}     %
\DeclareUnicodeCharacter{03A7}{X}                     %
\DeclareUnicodeCharacter{03A8}{\ensuremath{\Psi}}     %
\DeclareUnicodeCharacter{03A9}{\ensuremath{\Omega}}   %

\DeclareUnicodeCharacter{2260}{\ensuremath{\neq}}     %
\DeclareUnicodeCharacter{2261}{\ensuremath{\equiv}}   %
\DeclareUnicodeCharacter{2248}{\ensuremath{\approx}}  %
\DeclareUnicodeCharacter{2243}{\ensuremath{\simeq}}   %
\DeclareUnicodeCharacter{2245}{\ensuremath{\cong}}    %
\DeclareUnicodeCharacter{223C}{\ensuremath{\sim}}     %
\DeclareUnicodeCharacter{221D}{\ensuremath{\propto}}  %
\DeclareUnicodeCharacter{2264}{\ensuremath{\leq}}     %
\DeclareUnicodeCharacter{2265}{\ensuremath{\geq}}     %
\DeclareUnicodeCharacter{2208}{\ensuremath{\in}}      %
\DeclareUnicodeCharacter{2209}{\ensuremath{\notin}}   %

\DeclareUnicodeCharacter{2282}{\ensuremath{\subset}}  %
\DeclareUnicodeCharacter{2283}{\ensuremath{\supset}}  %
\DeclareUnicodeCharacter{2286}{\ensuremath{\subseteq}}%
\DeclareUnicodeCharacter{2287}{\ensuremath{\supseteq}}%
\DeclareUnicodeCharacter{2229}{\ensuremath{\cap}}     %
\DeclareUnicodeCharacter{222A}{\ensuremath{\cup}}     %

\DeclareUnicodeCharacter{2227}{\ensuremath{\wedge}}   %
\DeclareUnicodeCharacter{2228}{\ensuremath{\vee}}     %
\DeclareUnicodeCharacter{22A5}{\ensuremath{\perp}}    %
\DeclareUnicodeCharacter{2234}{\ensuremath{\therefore}} %
\DeclareUnicodeCharacter{2235}{\ensuremath{\because}} %

\DeclareUnicodeCharacter{2192}{\ensuremath{\rightarrow}}      %
\DeclareUnicodeCharacter{2190}{\ensuremath{\leftarrow}}       %
\DeclareUnicodeCharacter{21D2}{\ensuremath{\Rightarrow}}      %
\DeclareUnicodeCharacter{21D0}{\ensuremath{\Leftarrow}}       %
\DeclareUnicodeCharacter{21D4}{\ensuremath{\Leftrightarrow}}  %
\DeclareUnicodeCharacter{21A6}{\ensuremath{\mapsto}}          %

\DeclareUnicodeCharacter{221E}{\ensuremath{\infty}}   %
\DeclareUnicodeCharacter{2202}{\ensuremath{\partial}} %
\DeclareUnicodeCharacter{2207}{\ensuremath{\nabla}}   %
\DeclareUnicodeCharacter{220F}{\ensuremath{\prod}}    %
\DeclareUnicodeCharacter{2211}{\ensuremath{\sum}}     %
\DeclareUnicodeCharacter{222B}{\ensuremath{\int}}     %
\DeclareUnicodeCharacter{2206}{\ensuremath{\Delta}}   %
\DeclareUnicodeCharacter{221A}{\ensuremath{\sqrt{}}}  %

\DeclareUnicodeCharacter{00B1}{\ensuremath{\pm}}      %
\DeclareUnicodeCharacter{00D7}{\ensuremath{\times}}   %
\DeclareUnicodeCharacter{00B7}{\ensuremath{\cdot}}    %
\DeclareUnicodeCharacter{00B0}{\ensuremath{^\circ}}   %

\DeclareUnicodeCharacter{2016}{\ensuremath{\Vert}}    %

\DeclareUnicodeCharacter{2026}{\dots}                 %
\DeclareUnicodeCharacter{2032}{\ensuremath{'}}        %
\DeclareUnicodeCharacter{2033}{\ensuremath{''}}       %

\DeclareUnicodeCharacter{2200}{\ensuremath{\forall}}  %
\DeclareUnicodeCharacter{2203}{\ensuremath{\exists}}  %

\DeclareUnicodeCharacter{00B2}{\textsuperscript{2}}   %
\DeclareUnicodeCharacter{00B3}{\textsuperscript{3}}   %

\DeclareUnicodeCharacter{2212}{\ensuremath{-}}

\usepackage[capitalize,noabbrev]{cleveref}

\theoremstyle{plain}
\newtheorem{theorem}{Theorem}[section]

\theoremstyle{definition}
\newtheorem{definition}[theorem]{Definition}

\theoremstyle{remark}

\usepackage[textsize=tiny]{todonotes}

\title{Parason: Revealing Subtask- and Trial Parallelism \\ in LLM Reasoning}

\author{%
  Zhengyang Zhang$^{1}$ \quad 
  Zijian Zhang$^{1}$ $^{4}$ \quad
  Jiaxuan Gao$^{1}$ \quad \\
  \bf Shusheng Xu$^{2}$ \quad
  \bf Yi Wu$^{1}$ \quad
  \bf Song Han$^{3}$ $^{4}$ \quad
  \bf Ligeng Zhu$^{4}$ \thanks{ \texttt{ligengz@nvidia.com}} \\[1ex]
  $^{1}$ Tsinghua University \quad $^{2}$Independent Researcher \quad $^{3}$Massachusetts Institute of Technology \quad $^{4}$ NVIDIA \\
}

\begin{document}

\maketitle

\begin{abstract}
    Scaling test-time reasoning has substantially improved the problem-solving ability of large language models (LLMs), but standard autoregressive decoding still executes long reasoning traces sequentially, creating severe latency for difficult tasks (up to days and weeks). Parallel reasoning offers a natural remedy. However, prior systems primarily focus on \textit{Subtask Parallelism}, where the model learns to decompose a high-level task into smaller chunks that can be solved independently. This approach overlooks another pervasive form of parallelism: \textit{Trial Parallelism}, where multiple speculative attempts explore, verify, and aggregate competing hypotheses in parallel. In this paper, we introduce \textbf{Parason}, which reveals and learns both forms of parallelism in LLM reasoning. Our analysis identifies Trial Parallelism as the majority of parallelizable reasoning computation (65.5\% in DeepSeek-V4's reasoning steps in HLE), and it becomes increasingly dominant on hard problems. Guided by this taxonomy, Parason converts sequential reasoning traces into structured parallel trajectories with a context-free grammar, then trains models with Parallelism-Aware Group Relative Policy Optimization (PA-GRPO), whose reward jointly balances accuracy, latency, and the two parallelism ratios. At inference time, Parason executes the learned parallel structure through tool calls, translating theoretical savings to real-world wall-clock acceleration. Experiments on mathematical reasoning benchmarks including AIME24 and AIME25 show that Parason achieves \textbf{an average acceleration about 1.7$\times$} while maintaining competitive accuracy.
\end{abstract}
\begin{center}
\small
\href{https://zhengyangzhang06.github.io/parason-web/}{\textbf{Website}}
\quad\textbar\quad
\href{https://github.com/ZhengyangZhang06/parason-public}{\textbf{Code}}
\quad\textbar\quad
\href{https://huggingface.co/datasets/parallel-reasoner/parason/tree/main}{\textbf{Dataset}}
\end{center}

\begin{figure*}
    \centering
    \includegraphics[width=0.95\linewidth]{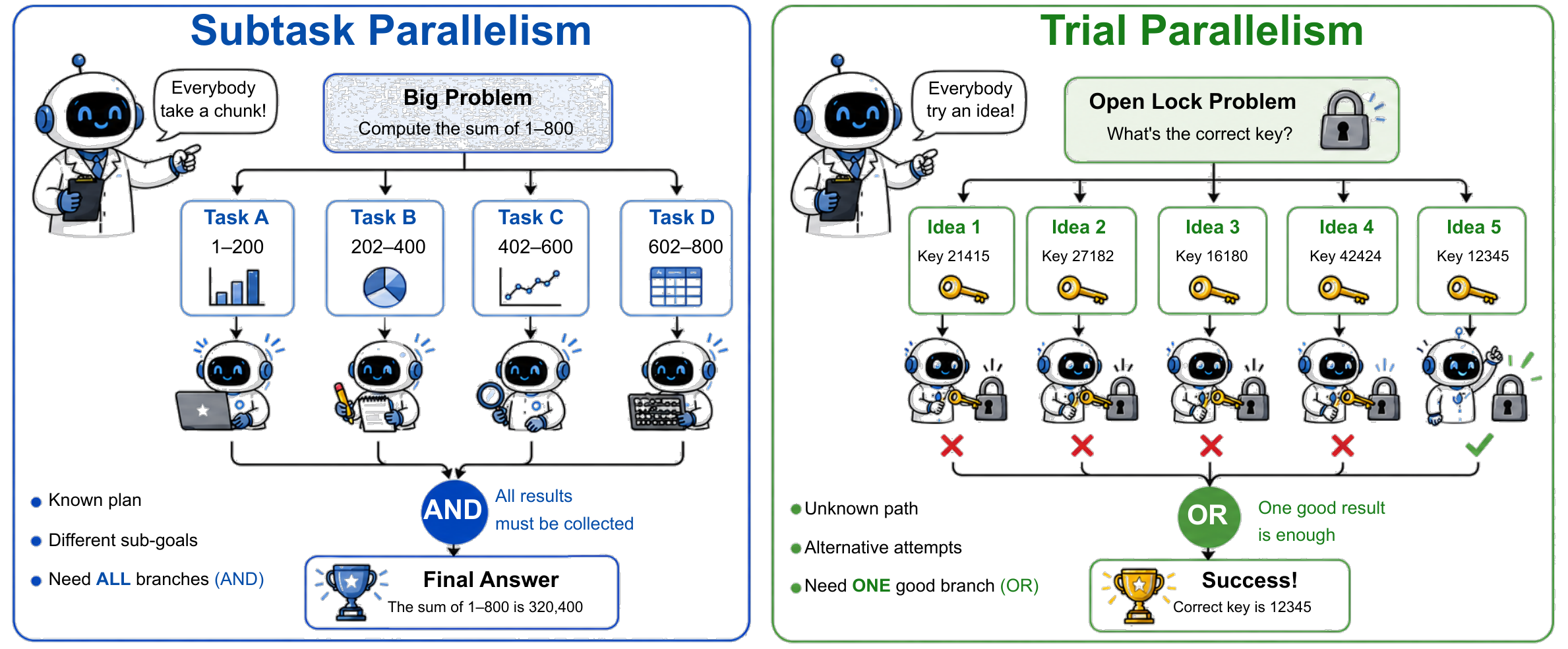}
    \caption{
        \textbf{Overview of two types of parallelism.}
        The figure contrasts Subtask Parallelism, where all decomposed branches are needed for the final answer, with Trial Parallelism, where multiple uncertain paths are explored as alternative (OR) branches. Despite this OR relation, the outputs of all Trial branches are concatenated into the subsequent context, making every branch's content available for final synthesis.
        While previous work mainly focuses on Subtask Parallelism, our analysis shows that Trial Parallelism accounts for the majority of parallelizable reasoning on HLE for every model we study, and exceeds 58\% for DeepSeek-R1 and DeepSeek-V4 across both datasets.
    }
    \label{fig:teaser_two_parallelism}
\end{figure*}

\section{Introduction}

Recent advances in large language model (LLM) reasoning have made a simple recipe increasingly effective: spend more test-time compute, and obtain better answers. Chain-of-Thought prompting~\citep{wei2022chain} and reasoning-oriented systems such as OpenAI o1~\citep{openai2024o1} and DeepSeek-R1~\citep{deepseekai2025deepseekr1incentivizingreasoningcapability} show that long, deliberate reasoning traces can substantially improve performance on mathematical and logical tasks. Yet this recipe has an immediate systems bottleneck. Standard autoregressive decoding serializes every token and every intermediate thought, so stronger reasoning often translates directly into longer waiting time. On difficult problems, reasoning traces can grow to hundreds of thousands or even millions of tokens~\citep{imo24,imojourney}, making the usual ``think longer'' strategy impractical for interactive agents, coding assistants, and other latency-sensitive applications. For example, Google's AlphaProof~\citep{alphaproof2024google} requires up to \textit{three days} to solve challenging problems from the International Mathematical Olympiad (IMO) 2024, highlighting the extreme latency inherent in current sequential reasoning systems.

A natural response is to make reasoning parallel: let multiple reasoning branches proceed concurrently instead of decoding every thought in sequence. Existing methods explore this direction through independent sampling, such as self-consistency and Best-of-N~\citep{wang2022self, pareco}, or adaptive branching frameworks~\citep{yang2025multiverse, jin2025learning, zheng2025parallelr1, threadweaver2025, wang2025survey}. However, methods such as Multiverse~\citep{yang2025multiverse}, ThreadWeaver~\citep{threadweaver2025}, and APR~\citep{pan2025learning} mainly exploit \emph{subtask-style} parallelism: splitting a high-level task into smaller chunks and merging their results. Trial-style exploration, which tries uncertain paths and incorporates diverse trajectories, is less discussed.

In this work, we argue that the key missing piece is a semantic taxonomy of parallel reasoning. As illustrated in Figure~\ref{fig:teaser_two_parallelism}, we identify two complementary forms of parallelism. \textbf{Subtask Parallelism} applies when a problem can be decomposed into independent steps: each branch solves a distinct sub-goal, and the final answer combines all branch results. This is the form emphasized by most prior adaptive parallel systems. \textbf{Trial Parallelism} applies when the model is uncertain about which path will work: multiple branches test competing hypotheses, and the final trajectory keeps the useful results. These modes have different execution semantics. Subtask branches are usually all necessary, while trial branches are speculative and most useful when the reasoning path is uncertain. Our empirical analysis shows that this distinction is practical, not merely conceptual. Trial Parallelism accounts for the majority of analyzed parallelizable steps on Humanity's Last Exam for every model we study, including \ul{73.8\% and 65.5\% for DeepSeek-R1 and DeepSeek-V4}, respectively. On OpenMath, it remains the majority for most models and reaches 68.5\% and 58.1\% for the same two open-source models. This suggests that hard reasoning is not only decomposition, but also trying and refining uncertain solution paths. Systems that focus solely on subtask decomposition therefore miss much of the computation in hard cases.

Guided by this observation, we introduce \textbf{Parason}, a training-and-inference codesign framework for revealing and exploiting both Subtask Parallelism and Trial Parallelism. Parason converts sequential reasoning traces into structured parallel trajectories without changing their final semantics, and uses a context-free grammar to mark parallel regions, individual branches, and summaries in an engine-parseable format. We then train the model with {Parallelism-Aware Group Relative Policy Optimization (PA-GRPO)}, which augments outcome-based reinforcement learning with rewards for lower latency along the longest token path and balanced use of the two parallelism types. At inference time, the generated structure is executed through tool calls, allowing Parason to dispatch independent workers with subtasks and competing trial branches. Across mathematical reasoning benchmarks including AIME24 and AIME25, Parason preserves competitive accuracy while reducing effective reasoning latency.
Our contributions are summarized as follows:
\begin{itemize}
    \item \textbf{A taxonomy of parallel reasoning.} We distinguish \textbf{Subtask Parallelism} and \textbf{Trial Parallelism}, identifying Trial Parallelism as the missing component. Based on this taxonomy, we introduce Parason, which converts sequential reasoning data into grammar-constrained parallel trajectories that can be parsed and executed by inference engines.
    \item \textbf{Parallelism-aware reinforcement learning.} We propose PA-GRPO, a multi-objective RL objective that jointly optimizes answer accuracy, latency along the longest token path, and the model's use of the two parallelism modes.
    \item \textbf{Empirical speedups with competitive accuracy.} Experiments across challenging math benchmarks demonstrate that Parason improves the latency--accuracy Pareto frontier, achieving an average acceleration about $1.7\times$ while maintaining competitive accuracy. Under latency-constrained settings, Parason matches the performance of an 8k-token latency budget using only 25\% of the budget.
    \item \textbf{Seamless inference engine support.} We define a CFG that marks parallel regions as tool calls and integrate Parason into SGLang~\citep{zheng2023sglang}. This makes parallel reasoning directly executable in a real inference engine, while prior work either reports only theoretical speedups that are hard to translate into actual latency reduction, or introduces complicated designs that require deep modifications to modern inference engines.
\end{itemize} %

\begin{figure*}[t]
\begin{tcolorbox}[
    enhanced,
    colback=white,
    colbacktitle=black,
    title={Sequential Reasoning is Naturally Parallelizable -- A Case of 24 Puzzle Solving},
    fontupper=\tiny, 
    attach boxed title to top left={yshift=-2mm, xshift=2mm},
    boxed title style={sharp corners, boxrule=0pt},
    arc=2pt,
    drop shadow={black!50!white},
    left=8pt, right=8pt, top=8pt, bottom=8pt 
]
\textbf{Prompt:} Calculate 24 using 6 9 9 10
\begin{Verbatim}[commandchars=\\\{\}, breaklines=true,breakautoindent=true,breaksymbol={}]
<think> We are given the numbers 6, 9, 9, 10 and we need to make 24 using basic arithmetic operations (+, -, *, /) and possibly parentheses. We need to find an expression that equals 24. Let's try different combinations: We have two 9's, one 6, and one 10. Goal: 24. Some common approaches: 24 = 4*6, 24=30-6, 24=2.4*10, etc.

\textcolor{blue}{Since we have a 6, maybe we can get 4 from 9,9,10 to multiply by 6? } \textcolor{green!45!black}{4*6=24. How to get 4 from 9,9,10? 10 - 9 = 1, then 9 - 1 = 8, not 4. Or (9+9)/10? =18/10=1.8, not 4. Maybe 24 = 30 - 6. So if we can make 30 from 9,9,10 and then subtract 6? But we have to use all numbers. So maybe (9+9+10) = 28, then 28 - 4? But we have 6. Alternatively, (9*10)/? Not sure. Or 24 = (10-6)*6, but we only have one 6. Maybe using fractions: 24 = 9/(something) etc. Let's try other combinations systematically: We can think of possible expressions: One idea: 10 * (9 - 6) = 10*3=30, then 30 - 9 = 21, not 24. 10 * (9 - 6) - 9 = 30-9=21. 9 * (10 - 6) = 9*4=36, then 36 - 9 = 27, not 24. (9+9)* (10-6)? =18*4=72 too big. (9+9-10)*6 = (8)*6=48, too big.}

\textcolor{blue}{Maybe divide: (9*10)/ (something) = 90/ something.} \textcolor{green!45!black}{90/4 = 22.5, not 24. Or (9*10*9)/ something. 810/33.75 = 24? Not likely.} \textcolor{blue}{Maybe using subtraction and division: 24 = (10*9)/(something) - something.} \textcolor{green!45!black}{Let's try: 9*10 = 90, 90/6=15, then 15+9=24! That works: 9*10/6+9 = 90/6+9=15+9=24. But we have to use two 9's and one 10 and one 6. In that expression we used one 9, one 10, one 6, and then added the other 9. So it's 9*10/6+9. But check order of operations: multiplication and division first: 9*10=90, 90/6=15, 15+9=24. Yes.}

So the expression is: 9 × 10 ÷ 6 + 9 = 24. we found a solution. \textcolor{blue}{But let's check if there are other ways: maybe (9 - 6) × (10 - √9)} but we cannot use square root unless specified. So basic operations only. \textcolor{blue}{Another: (10 - 9/9)*? (10-1)=9, 9*?} \textcolor{green!45!black}{Not 24. (9+9+10-6)? =22, not 24.(9*9-10)/6? (81-10)/6=71/6≈11.83, no.}

So the one we found seems simplest. Thus, the answer is: 9 × 10 ÷ 6 + 9 = 24. </think>
\end{Verbatim}
One solution is: 9 × 10 ÷ 6 + 9 = 24
\end{tcolorbox}
\caption{
    \textbf{Visualization of Trial and Subtask Parallelism.} The 24 Puzzle trace generated by DeepSeek-R1~\citep{deepseekai2025deepseekr1incentivizingreasoningcapability} highlights \textcolor{blue}{Subtask Parallelism in blue} and \textcolor{green!45!black}{Trial Parallelism in green}. Subtask branches implement divide-and-conquer execution, while Trial branches explore and verify competing hypotheses; \textit{Trial steps occupy the majority} of reasoning tokens.
}
\label{fig:dpsk_reasoning_example}
\end{figure*}

\section{Related Work}

\paragraph{Test-time scaling for LLM reasoning.}
Scaling inference-time computation has become a central recipe for improving LLM reasoning. CoT prompting~\citep{wei2022chain} elicits intermediate derivations, while recent reasoning models such as OpenAI o1~\citep{openai2024o1} and DeepSeek-R1~\citep{deepseekai2025deepseekr1incentivizingreasoningcapability} further improve performance by producing longer and more deliberate reasoning traces. This sequential scaling is powerful but expensive: autoregressive decoding generates every token one after another, so longer reasoning directly increases latency. Parallel reasoning is an attractive direction for preserving test-time compute while shortening the longest token path.

\paragraph{Independent parallel sampling and majority voting.}
A simple form of parallel reasoning is to sample multiple complete solutions and aggregate their answers. Self-consistency~\citep{wang2022self}, verifier-guided Best-of-N sampling~\citep{cobbe2021trainingverifierssolvemath}, and confidence-based variants such as DeepConf~\citep{fu2025deep} improve robustness by exploring several candidate trajectories. These methods implicitly use trial-style computation, since different samples may try different solution paths. However, these samples are independent full traces. They do not share intermediate work, expose branch structure inside a trace, or tell an inference engine where parallel execution should begin and end. As a result, they improve accuracy at the cost of redundant computation with little control over latency.

\paragraph{Structured and adaptive parallel reasoning.}
Another line of work introduces explicit structures for parallel reasoning. Tree-of-Thoughts~\citep{yao2023tree}, Graph-of-Thoughts~\citep{besta2023graph}, Skeleton-of-Thought~\citep{ning2023skeleton}, and agentic decomposition methods split reasoning into trees, graphs, outlines, or sub-agents. More recent adaptive systems, including PASTA~\citep{jin2025learning}, Multiverse~\citep{yang2025multiverse}, Parallel-R1~\citep{zheng2025parallelr1}, APR~\citep{pan2025learning}, ThreadWeaver~\citep{threadweaver2025}, and PaCoRe~\citep{pareco}, train or prompt models to create parallel branches and merge their outputs. PaCoRe further scales test-time compute with multiple rounds of coordinated parallel exploration and message passing. Figure~\ref{fig:dpsk_reasoning_example} visualizes the same broader opportunity: \textit{LLM reasoning can be parallelized}.

\begin{table}[t]
    \centering
    \small
    \begin{tabular}{l c c c c}
        \toprule
        \multirow{2}{*}{Model} & \multicolumn{2}{c}{Dataset: OpenMath} & \multicolumn{2}{c}{Dataset: HLE} \\
        \cmidrule(lr){2-3} \cmidrule(lr){4-5}
        & Subtask (\%) & Trial (\%) & Subtask (\%) & Trial (\%) \\
        \midrule
        \rowcolor{black!10} Open Source Models & & & &  \\
        DeepSeek-R1~\citep{deepseekai2025deepseekr1incentivizingreasoningcapability} & 31.5 & $68.5_{\scriptscriptstyle\textcolor{green!50!black}{+37.00}}$ & 26.2 & $73.8_{\scriptscriptstyle\textcolor{green!50!black}{+47.60}}$ \\
        Qwen3-30B-A3B~\citep{yang2025qwen3} & 43.6 & $56.3_{\scriptscriptstyle\textcolor{green!50!black}{+12.70}}$ & 38.8 & $61.2_{\scriptscriptstyle\textcolor{green!50!black}{+22.40}}$ \\
        MiniMax 2.7~\citep{minimax2025m2} & 47.7 & $52.3_{\scriptscriptstyle\textcolor{green!50!black}{+4.60}}$ & 38.2 & $61.8_{\scriptscriptstyle\textcolor{green!50!black}{+23.60}}$ \\
        Kimi-2.6~\citep{moonshot2025kimi} & 43.0 & $57.0_{\scriptscriptstyle\textcolor{green!50!black}{+14.00}}$ & 42.1 & $57.9_{\scriptscriptstyle\textcolor{green!50!black}{+15.80}}$ \\
        DeepSeek-V4~\citep{deepseekai2026deepseekv4} & 41.9 & $58.1_{\scriptscriptstyle\textcolor{green!50!black}{+16.20}}$ & 34.5 & $65.5_{\scriptscriptstyle\textcolor{green!50!black}{+31.00}}$ \\
        \midrule
        \rowcolor{black!10} Commercial Models & & & &  \\
        Gemini-2.5-Pro~\citep{google-gemini-2.5} & 62.9 & $37.1_{\scriptscriptstyle\textcolor{red}{-25.80}}$ & 49.4 & $50.6_{\scriptscriptstyle\textcolor{green!50!black}{+1.20}}$ \\
        Claude-Opus-4.5~\citep{anthropic2025claudeopus45} & 44.9 & $55.1_{\scriptscriptstyle\textcolor{green!50!black}{+10.20}}$ & 29.9 & $70.1_{\scriptscriptstyle\textcolor{green!50!black}{+40.20}}$ \\
        Gemini-3-Pro~\citep{google2025gemini3pro} & 57.4 & $42.6_{\scriptscriptstyle\textcolor{red}{-14.80}}$ & 46.9 & $53.1_{\scriptscriptstyle\textcolor{green!50!black}{+6.20}}$ \\
        GPT-5.5~\citep{openai2026gpt55} & 34.4& $65.6_{\scriptscriptstyle\textcolor{green!50!black}{+31.20}}$& 23.9 & $76.1_{\scriptscriptstyle\textcolor{green!50!black}{+52.20}}$ \\
        \bottomrule
    \end{tabular}
    \caption{\textbf{Trial Parallelism dominates on harder reasoning datasets.} We report Subtask and Trial ratios in reasoning traces from OpenMathReasoning~\citep{openmath} and Humanity's Last Exam (HLE)~\citep{hle25}. Ratios are computed on the first 100 problems of each dataset, and annotation details are provided in Appendix~A. For closed-source models, the original thinking trace is hidden, thus we counted summary statistics instead.}
    \label{tab:reasoning_composition_combined}
\end{table}

Yet their main mechanism is still \textit{subtask-centric}: the model decomposes a high-level task into smaller chunks, runs the chunks in parallel, and merges the results. This is effective when the problem admits a clean divide-and-conquer structure, but it misses a common behavior in difficult reasoning: trying uncertain paths, rejecting failed attempts, and keeping the path that works. Our analysis shows that this missing \textit{Trial Parallelism is not rare}; it accounts for most parallelizable reasoning steps in hard cases. Parason therefore separates \textbf{Subtask Parallelism} from \textbf{Trial Parallelism} and trains the model to exploit both forms explicitly.

\section{Methodology}
\label{sec:method}

Parason turns long sequential reasoning traces into executable parallel programs, which is critical for long-horizon tasks such as Google's AlphaProof~\citep{alphaproof2024google}, OpenAI's IMO reasoning efforts~\citep{imojourney}, and recent agentic coding workflows such as Claude Code~\citep{anthropic2025claudecode} and OpenAI Codex~\citep{openai2025codex}. The framework has three parts. First, we define two semantic forms of parallel reasoning: Subtask Parallelism decomposes a problem into independent required parts, while Trial Parallelism explores competing uncertain attempts and incorporates the diverse exploration into the final reasoning path. Second, we define a context-free grammar (CFG) that gives Parason a formal syntax, helps data generation, processing, and rollouts avoid syntax errors, and makes later system integration simple. Third, we propose Parallelism-Aware GRPO (PA-GRPO), a training algorithm that teaches the model to use parallelism effectively while preserving final-answer correctness.

\subsection{Parallelism in Reasoning}
\label{sec:taxonomy}

A reasoning trace is not uniformly sequential, as shown by the visualization of Trial and Subtask Parallelism in Figure~\ref{fig:dpsk_reasoning_example}. Some steps derive independent facts that will all be used later; other steps try uncertain ideas, explore multiple paths, and aggregate their findings. Following the overview and examples in Figures~\ref{fig:teaser_two_parallelism}, \ref{fig:dpsk_reasoning_example}, and~\ref{fig:dpsk_reasoning_example_cfg}, Parason treats these behaviors as two semantic forms of parallelism with different merge rules.

\textbf{Subtask Parallelism} is an \emph{AND-branch} form of parallelism. A problem is split into independent sub-goals $g_1,\ldots,g_k$, each branch computes a necessary result $r_i$, and the main trajectory aggregates all results $\{r_i\}_{i=1}^k$ to continue the derivation. These branches are parallel because they do not depend on one another, but all of them are mandatory: dropping one branch would remove the information required for generating the final answer. For example, a geometry solution may compute two distances independently before combining them in a final formula. This is the dominant form assumed by prior adaptive parallel systems~\citep{yang2025multiverse,threadweaver2025,pan2025learning,pareco}.

\textbf{Trial Parallelism} is an \emph{OR-branch} form of parallelism. Given an uncertain state, the model launches competing attempts $a_1,\ldots,a_k$ that test different hypotheses, heuristics, or solution routes. The merge step concatenates the branches to incorporate the exploratory trajectories into the reasoning history, rather than selecting a single branch. Thus, Trial Parallelism is parallel search inside a single reasoning trace, rather than decomposition into mandatory parts. This mirrors how strong reasoning models often behave on hard problems: they try a path, detect a dead end, and revise. Our empirical analysis in Table~\ref{tab:reasoning_composition_combined} shows that Trial Parallelism accounts for more than 50\% of parallelizable steps for every model on HLE~\citep{hle25} and for most models on OpenMath, with higher Trial shares on HLE across all listed models.

\begin{figure*}[t]
\begin{tcolorbox}[
    enhanced,
    colback=white,
    colbacktitle=black,
    title={Parallel Reasoning in Parason -- A Case of 24 Puzzle Solving},
    fontupper=\tiny,
    attach boxed title to top left={yshift=-2mm, xshift=2mm},
    boxed title style={sharp corners, boxrule=0pt},
    arc=2pt,
    drop shadow={black!50!white},
    left=8pt, right=8pt, top=8pt, bottom=8pt
]
\textbf{Prompt:} Calculate 24 using 6, 9, 9, 10.
\begin{Verbatim}[commandchars=\\\{\}, breaklines=true, breakautoindent=true, breaksymbol={}]
<think> We are given the numbers 6, 9, 9, 10 and need to make 24 using basic arithmetic operations (+, -, *, /). Goal: 24.

<Parallel>
 <Outlines>
  \textcolor{blue}{<Subtask> 1. Try to use 6 as a multiplier and derive 4 from 9, 9, 10. </Subtask>}
  \textcolor{blue}{<Subtask> 2. Try division with 9*10 and use the remaining 6 and 9. </Subtask>}
  \textcolor{blue}{<Subtask> 3. Verify whether there are simpler alternatives. </Subtask>}
\end{Verbatim}
\begin{minipage}[t]{0.32\linewidth}
\begin{Verbatim}[commandchars=\\\{\}, breaklines=true, breakautoindent=true, breaksymbol={}]
 <Thread> 1.
  <Parallel>
   <Outlines>
    \textcolor{green!45!black}{<Trial> 1. Try use 9, 9, 10 to make 4. 10-9=1, then 9-1=8, not 4. This route does not find 4. </Trial>}
    \textcolor{green!45!black}{<Trial> 2. Try 10*(9-6)-9. This gives 21, not 24. This route does not find 24. </Trial>}
    \textcolor{green!45!black}{<Trial> 3. Try 9*(10-6)-9. This gives 27, not 24. </Trial>}
    \textcolor{green!45!black}{<Trial> 4. Try (9+9)*(10-6). This gives 72, not 24. </Trial>}
   </Outlines>
  </Parallel>
 </Thread>
\end{Verbatim}
\end{minipage}
\hfill
\begin{minipage}[t]{0.32\linewidth}
\begin{Verbatim}[commandchars=\\\{\}, breaklines=true, breakautoindent=true, breaksymbol={}]
 <Thread> 2.
  <Parallel>
   <Outlines>
    \textcolor{green!45!black}{<Trial> 1. Try 9*10/6+9. 90/6=15, 15+9=24. This works. </Trial>}
   </Outlines>
  </Parallel>
 </Thread>
\end{Verbatim}
\end{minipage}
\hfill
\begin{minipage}[t]{0.32\linewidth}
\begin{Verbatim}[commandchars=\\\{\}, breaklines=true, breakautoindent=true, breaksymbol={}]
 <Thread> 3.
  <Parallel>
   <Outlines>
    \textcolor{green!45!black}{<Trial> 1. Try (9-6)*(10-sqrt(9)). Square root is not allowed. </Trial>}
    \textcolor{green!45!black}{<Trial> 2. Try 9+9+10-6. Result is 22, not 24. </Trial>}
    \textcolor{green!45!black}{<Trial> 3. Try (9*9-10)/6. Result is 11.83, not 24. </Trial>}
   </Outlines>
  </Parallel>
 </Thread>
</Outlines>
\end{Verbatim}
\end{minipage}
\begin{Verbatim}[commandchars=\\\{\}, breaklines=true, breakautoindent=true, breaksymbol={}]
 </Outlines>
</Parallel>
Thus, the answer is 9 × 10 ÷ 6 + 9 = 24. </think>
\end{Verbatim}
\end{tcolorbox}
\caption{
    \textbf{Solving the 24 Puzzle in Parason format.} This example is adapted from a real 24 Puzzle reasoning trace. The \textcolor{blue}{blue} region gives the Subtask plan, splitting the problem into three independent subtasks. The \textcolor{green!45!black}{green} regions show Trial Parallelism, where each subtask explores candidate routes in parallel. The correct solution appears in the first Trial branch of the second subtask.
}
\label{fig:dpsk_reasoning_example_cfg}
\end{figure*}

\begin{figure}[t]
\begin{tcolorbox}[
    enhanced,
    colback=white,
    colbacktitle=black,
    title={Context-Free Grammar (CFG) for Parason},
    fontupper=\footnotesize, 
    attach boxed title to top left={yshift=-2mm, xshift=2mm},
    boxed title style={sharp corners, boxrule=0pt},
    arc=2pt,
    drop shadow={black!50!white},
    left=8pt, right=8pt, top=8pt, bottom=8pt 
]
Let $G = (V, \Sigma, R, S)$ be the context-free grammar for structured parallel reasoning.
\begin{minipage}[t]{0.43\linewidth}
\textbf{Terminology}
\begin{itemize}
    \setlength\itemsep{1pt}
    \setlength\parsep{0pt}
    \item \textbf{Terminals ($\Sigma$):}
    \{\texttt{<think>}, \texttt{<Parallel>}, \texttt{<Outlines>}, \texttt{<Subtask>}, \texttt{<Trial>}, \texttt{<Thread>}, their matching closing tags, \texttt{Text}\}
    \item \textbf{Non-terminals ($V$):}
    \begin{itemize}
        \setlength\itemsep{0pt}
        \setlength\parsep{0pt}
        \item $S$: start of reasoning
        \item $B$: reasoning body
        \item $P$: parallel region
        \item $O$: outlines block
        \item $G$: list of outline entries
        \item $E$: one Subtask or Trial entry
        \item $H$: list of thread bodies
        \item $W$: one Thread body
        \item $T$: free-form reasoning text
    \end{itemize}
    \item \textbf{Start symbol:} $S$
    \item \textbf{Rules:} $R$ defines valid tag nesting and branch structure.
\end{itemize}
\end{minipage}
\hfill
\begin{minipage}[t]{0.54\linewidth}

\hspace*{10em}\textbf{Production Rules $R$}\par
\begin{align*}
    \texttt{S} &\rightarrow \text{\texttt{<think>}} \, \texttt{B} \, \text{\texttt{</think>}} \\
    \texttt{B} &\rightarrow \texttt{T} \, \texttt{B} \mid \texttt{P} \, \texttt{B} \mid \epsilon \\
    \texttt{P} &\rightarrow \text{\texttt{<Parallel>}} \, \texttt{O} \, \texttt{H} \, \text{\texttt{</Parallel>}} \\
    \texttt{O} &\rightarrow \text{\texttt{<Outlines>}} \, \texttt{G} \, \text{\texttt{</Outlines>}} \\
    \texttt{G} &\rightarrow \texttt{E} \, \texttt{G} \mid \texttt{E} \\
    \texttt{E} &\rightarrow \text{\texttt{<Subtask>}} \, \texttt{T} \, \text{\texttt{</Subtask>}} \\
    & \quad\mid \text{\texttt{<Trial>}} \, \texttt{T} \, \text{\texttt{</Trial>}} \\
    \texttt{H} &\rightarrow \texttt{W} \, \texttt{H} \mid \texttt{W} \\
    \texttt{W} &\rightarrow \text{\texttt{<Thread>}} \, \texttt{B} \, \text{\texttt{</Thread>}} \\
    \texttt{T} &\rightarrow \text{nonempty free-form text excluding reserved tags}
\end{align*}
\end{minipage}
\end{tcolorbox}
\caption{\textbf{Context-Free Grammar for Parason.} Each parallel region contains an Outlines block with one or more correctly matched Subtask or Trial entries, followed by one or more Thread bodies. As in Figure~\ref{fig:dpsk_reasoning_example_cfg}, each Thread may contain free-form reasoning or a nested parallel region.}
\label{fig:parason_cfg}
\end{figure}

\subsection{Parallel Tracjectory Format}
\label{sec:formal_def}

To make these two modes executable, Parason represents reasoning traces with a grammar-constrained format. We extend ordinary \texttt{<think>} traces with parallel tags: \texttt{<Parallel>} marks a parallel region, \texttt{<Outlines>} describes the purpose of the region, \texttt{<Subtask>} and \texttt{<Trial>} mark branch type, \texttt{<Thread>} stores branch outcome. The full CFG is shown in Figure~\ref{fig:parason_cfg}. While real-world reasoning often exhibits mixed dependencies (e.g., a subtask internally spawning trial branches), our training data keeps a strict separation for simplicity. However, we observe that the model can generalize to these complex compositions during evaluation.

The format is designed around two requirements. First, it is \emph{semantic}: the tags indicate whether a branch is a necessary subtask or a speculative trial branch, not merely that it can run in parallel. This matters because prior parallel-reasoning methods often mix the two forms under a single branching interface, making it hard to decide whether branches should be all merged or aggregated as exploration history. Second, it is \emph{engine-parseable}: each branch has explicit start and stop tags, so an inference runtime can dispatch workers without modifying the model architecture. Compared with free-form summaries or prompt-only, the CFG gives the training and inference infra a shared contract.

\subsection{Parallelism-Aware Reinforcement Learning}
\label{sec:rl}

Supervised conversion teaches the model to imitate parallel traces, but it does not directly optimize latency or decide \textit{when} and \textit{what} parallelism is worth using. We therefore RL fine-tune with \textbf{Parallelism-Aware Group Relative Policy Optimization (PA-GRPO)}. The reward keeps correctness as the primary objective, while shaping the model toward useful parallel branches. For a sampled trajectory $i$, let $T_i$ denote its token-level latency, and let $R_{\mathrm{subtask}}^i$ and $R_{\mathrm{trial}}^i$ be the ratios of tokens placed in Subtask and Trial branches. We define:

\begin{definition}
\textbf{Parallel RL Reward}
\begin{align*}
    R_i
&= -1 + 2 \times \mathbbm{1}(\mathrm{correct}_i) \\
&\quad \times
\Biggl(
1 - \alpha f\!\left(\frac{T_i - \mu_T}{\sigma_T}\right)
+ {\color{blue}\beta_{\text{subtask}}
    f\!\left(\frac{R_{\mathrm{subtask}}^i - \mu_s}{\sigma_s}\right)} \\
&\qquad
+ {\color{green!45!black}\beta_{\text{trial}}
    f\!\left(\frac{R_{\mathrm{trial}}^i - \mu_r}{\sigma_r}\right)}
+ \min\!\left(\rho \cdot \eta(s),\, \rho_{\mathrm{clip}}\right)
\Biggr).
\end{align*}
\label{eq:reward}
\end{definition}
The symbols in Eq.~\ref{eq:reward} are defined as follows:
\par\smallskip
\noindent
\begin{minipage}[t]{0.48\linewidth}
\footnotesize
\begin{itemize}
    \setlength\itemsep{1pt}
    \setlength\parsep{0pt}
    \item $R_i$: final reward for sampled trajectory $i$.
    \item $\mathbbm{1}(\mathrm{correct}_i)$: correctness indicator; incorrect answers receive the base negative reward.
    \item $T_i$: token-level latency of trajectory $i$, measured by its critical-path token count.
    \item $\alpha$: coefficient controlling the normalized latency penalty.
    \item {\color{blue}$R_{\mathrm{subtask}}^i$}: Number of tokens placed in Subtask Parallelism branches/Total generated tokens
    \item{$\eta(s)$}: $\eta(s) = 1 - \frac{L_{\text{latency}}}{L_{\text{total}}}$, where $L_{\text{latency}}$ and $L_{\text{total}}$ denote the number of latency tokens and the total number of tokens, respectively.
    \item {$\rho$, $\rho_{clip}$}: Parameters for controlling acceleration reward.
\end{itemize}
\end{minipage}
\hfill
\begin{minipage}[t]{0.48\linewidth}
\footnotesize
\begin{itemize}
    \setlength\itemsep{1pt}
    \setlength\parsep{0pt}
    \item {\color{green!45!black}$R_{\mathrm{trial}}^i$}: Number of tokens placed in Trial Parallelism branches/Total generated tokens
    \item {\color{blue}$\beta_{\text{subtask}}$} and {\color{green!45!black}$\beta_{\text{trial}}$}: incentives for Subtask and Trial Parallelism.
    \item $f$: shaping function applied to normalized reward signals; here we use the linear function $y=x$.
    \item $\mu_T,\mu_s,\mu_r$ and $\sigma_T,\sigma_s,\sigma_r$: means and standard deviations used to normalize latency, Subtask ratio, and Trial ratio, respectively.
\end{itemize}
\end{minipage}
\par\smallskip

The first term makes correctness the primary signal by assigning a negative base reward to incorrect answers and a positive base reward to correct answers. The latency term, weighted by $\alpha$, penalizes trajectories with high normalized critical-path latency. The Subtask and Trial incentives, weighted by $\beta_{\text{subtask}}$ and $\beta_{\text{trial}}$, encourage useful parallel structure, while the clipped acceleration reward directly favors reductions in the critical path relative to total generation length. It is important to note that $\beta_{\text{subtask}}$ and $\beta_{\text{trial}}$ target fundamentally different optimization goals: Trial Parallelism improves accuracy by trading total tokens for search width, while Subtask Parallelism reduces latency by compressing the longest token path. In practice, this reward lets us separately tune these two dimensions, as detailed in Section~\ref{sec:ablations_findings}.

\subsection{Integration into Modern Inference Engine}
\label{sec:inference_integration}

Parason integrates with modern inference engines by treating parallel branches as tool calls rather than requiring deep runtime changes. Following the CFG in Figure~\ref{fig:parason_cfg}, the runtime decodes normally until a \texttt{<Parallel>} region appears, parses the \texttt{<Outlines>}, dispatches each \texttt{<Subtask>} or \texttt{<Trial>} as an independent worker, and returns the resulting \texttt{<Thread>} outputs to the main trajectory; Figure~\ref{fig:dpsk_reasoning_example_cfg} shows this format on the 24 Puzzle. XGrammar~\citep{dong2024xgrammar} enforces valid tags, and we implement this tool-call execution in SGLang~\citep{zheng2023sglang}.

\begin{table}[t]
    \centering
    \footnotesize
    \setlength{\tabcolsep}{5.5pt}
    \renewcommand{\arraystretch}{1.15}
    \begin{tabular}{l c c c c c | c}
        \toprule
        \multicolumn{1}{c}{\multirow{1}{*}{Method}} & \multirow{1}{*}{AIME24} & \multirow{1}{*}{AIME25} & \multirow{1}{*}{Math500} & \multirow{1}{*}{AMC} & \multirow{1}{*}{Avg} & \makecell{Avg. Latency \\ \#Tokens} \\
        \midrule[0.8pt]
        Parallel-R1 (4B) & 19.4 & 19.2 & - & - & - & - \\
        ShorterBetter (7B) & 53.3 & - & - & - & - & - \\
        ThinkPrune (32B) & 72.5 & - & 93.8 & \textbf{95.9} & - & - \\
        DYNASOR-COT (32B) & \underline{78.0} & - & 93.0 & 94.0 & - & - \\
        Dynamic Early Exit (32B) & 70.0 & - & \textbf{94.8} & 95.0 & - & - \\
        AdaptThink (7B) & 55.6 & - & 92.0 & - & - & - \\
        Multiverse (32B) & 53.8 & 45.8 & 91.8 & - & - & - \\
        ThreadWeaver (8B) & \textbf{79.9} & 60.5 & 92.3 & 91.4 & 81.0 & 14.8k \\
        \midrule[0.8pt]
        Parason-8B (w/vanilla SFT) & 73.5 & 67.9 & 93.4 & 93.9 & 82.2 & 13.2k \\
        \quad + PA-GRPO ($\beta_{\text{subtask}} = 0.025$) & 75.1 & 69.7 & 93.8 & 96.6 & 83.8 & 13.8k \\
        \quad  + PA-GRPO ($\beta_{\text{subtask}} = 0.050$) & 76.3 & \underline{70.0} & \textbf{94.6} & 95.0 & 84.0 & 15.6k \\
        \quad  + PA-GRPO ($\beta_{\text{subtask}} = 0.100$) & \underline{77.3} & 68.5 & \textbf{94.6} & 96.3 & 84.2 & 14.1k \\
        \quad  + PA-GRPO ($\beta_{\text{trial}} = 0.025$)  & 75.5 & 69.9 & \underline{94.4} & 96.3 & 84.0 & 14.2k \\
        \quad  + PA-GRPO ($\beta_{\text{trial}} = 0.050$)  & 76.5 & \textbf{70.6} & 94.0 & 96.9 & \underline{84.5} & 14.5k \\
        \quad  + PA-GRPO ($\beta_{\text{trial}} = 0.100$)  & \textbf{78.2} & 68.9 & 94.2 & \textbf{97.5} & \textbf{84.7} & 14.5k \\
        \addlinespace[2pt]
        \multicolumn{7}{l}{\textit{Additional runs with $\alpha=0.1$}} \\
        \quad + PA-GRPO ($\beta_{\text{subtask}} = 0.025$) & 75.7 & 67.6 & 93.6 & 95.9 & 83.2 & \underline{12.2k} \\
        \quad  + PA-GRPO ($\beta_{\text{subtask}} = 0.050$) & 74.2 & 66.1 & 93.8 & 95.9 & 82.5 & \underline{12.2k} \\
        \quad  + PA-GRPO ($\beta_{\text{subtask}} = 0.100$) & 75.5 & 65.9 & 92.6 & \textbf{97.5} & 82.9 & 13.2k \\
        \quad  + PA-GRPO ($\beta_{\text{trial}} = 0.025$)  & 76.9 & 66.6 & \textbf{94.6} & \underline{97.2} & 83.8 & \textbf{12.1k} \\
        \quad  + PA-GRPO ($\beta_{\text{trial}} = 0.050$)  & \underline{77.3} & 67.3 & 93.4 & 96.6 & 83.7 & 12.5k \\
        \quad  + PA-GRPO ($\beta_{\text{trial}} = 0.100$)  & \textbf{78.2} & 69.2 & 93.4 & 95.6 & 84.1 & 13.4k \\
        \bottomrule
    \end{tabular}
    \captionof{table}{\textbf{Trial rewards improve accuracy, while Subtask rewards reduce latency.} We compare Parason with prior systems and ablate $\beta_{\text{subtask}}$ and $\beta_{\text{trial}}$. Avg and Token Latency average AIME24, AIME25, Math500, and AMC for each PA-GRPO row. Bold and underlined values mark the highest and second-highest PA-GRPO results, except Token Latency, where lower is better.}
    \label{tab:comparison_with_baselines}
\end{table}

\section{Experiments}
\label{sec:experiments}
\label{sec:experiment_setup}

We provide details on training data curation, training framework, and evaluation below, and will open-source our implementation to facilitate reproducibility.

\paragraph{Training Data Curation.}
\label{sec:data_curation}
We build parallel training data from the 964 annotated Qwen3-8B traces released by ThreadWeaver~\citep{threadweaver2025}. Gemini-3-Flash labels each parallel stage as Subtask or Trial; for Trial stages, Qwen3-8B samples extra branches and Gemini-3-Flash generates their \texttt{<Trial>} goals. For reinforcement learning (RL), we use Polaris-53k~\citep{Polaris2025}, a collection of about 53,000 complex reasoning problems.

\paragraph{Training Framework.}
We use a two-stage post-training pipeline. First, we fine-tune the model with TRL~\citep{vonwerra2020trl} on the curated parallel trajectories. Second, we train with VeRL~\citep{sheng2024hybridflow} using Parallelism-Aware GRPO (PA-GRPO). Unless otherwise noted, RL uses learning rate $1\mathrm{e}{-6}$, batch size 128, 8 rollouts, lower clip ratio 0.2, and upper clip ratio 0.28; SFT uses learning rate $1\mathrm{e}{-5}$, batch size 16, 8 epochs, and a cosine learning-rate scheduler. 

\paragraph{Evaluation.}
We evaluate on AIME 2024~\citep{aime2024}, AIME 2025~\citep{aime2025}, AMC~\citep{amc2023}, Math500~\citep{math500}, and Minerva Math~\citep{lewkowycz2022minerva}, and compare against sequential baselines trained from the same base model and data source as well as external parallel and efficient-reasoning systems. We report final-answer accuracy and executable latency. For Parason, \textbf{model latency} is the token length of the longest generation path: sequential tokens are counted normally, while each \texttt{<Parallel>} block contributes the maximum branch length rather than the sum of all branches. This metric captures the wall-clock benefit available to an inference engine that runs branches concurrently. 

\subsection{Main Results}
\label{sec:main_results}

\begin{table*}[t]
    \centering
    \setlength{\tabcolsep}{10pt}
    \renewcommand{\arraystretch}{1.15}
    \small
    \begin{tabular}{l c c c c}
        \toprule
        & \multicolumn{4}{c}{AIME24 Accuracy (\%)} \\
        \cmidrule(lr){2-5}
        Model & $B=2{,}048$ & $B=8{,}192$ & $B=16{,}384$ & $B=24{,}576$ \\
        \midrule
        SFT only & 16.8 & 41.8 & 66.8 & 70.2 \\
        PA-GRPO ($\beta_{\text{subtask}}=0.025$) & $16.7_{\scriptscriptstyle\textcolor{red}{-0.10}}$ & $45.2_{\scriptscriptstyle\textcolor{green!50!black}{+3.40}}$ & $68.0_{\scriptscriptstyle\textcolor{green!50!black}{+1.20}}$ & $73.2_{\scriptscriptstyle\textcolor{green!50!black}{+3.00}}$ \\
        PA-GRPO ($\beta_{\text{subtask}}=0.050$) & $26.8_{\scriptscriptstyle\textcolor{green!50!black}{+10.00}}$ & $60.0_{\scriptscriptstyle\textcolor{green!50!black}{+18.20}}$ & $72.2_{\scriptscriptstyle\textcolor{green!50!black}{+5.40}}$ & $74.8_{\scriptscriptstyle\textcolor{green!50!black}{+4.60}}$ \\
        PA-GRPO ($\beta_{\text{subtask}}=0.100$) & \textbf{34.7}$_{\scriptscriptstyle\textcolor{green!50!black}{+17.90}}$ & \textbf{60.3}$_{\scriptscriptstyle\textcolor{green!50!black}{+18.50}}$ & \textbf{74.4}$_{\scriptscriptstyle\textcolor{green!50!black}{+7.60}}$ & \textbf{75.7}$_{\scriptscriptstyle\textcolor{green!50!black}{+5.50}}$ \\
        \bottomrule
    \end{tabular}
    \caption{\textbf{Accuracy under token-latency budgets.} Each column reports AIME24 accuracy for a fixed thinking budget $B$ on the longest token path. Subscripts show absolute accuracy-point gains relative to SFT only under the same budget; red indicates a decrease. Subtask-aware PA-GRPO provides the largest gains in low-budget regimes: at $B=2{,}048$, $\beta_{\text{subtask}}=0.100$ reaches 34.7\% accuracy, compared with 16.8\% for SFT only.}
    \vspace{-10pt}
    \label{tab:main-comparison}
\end{table*}

\textbf{Parason improves accuracy over prior parallel reasoning systems.} Table~\ref{tab:comparison_with_baselines} compares Parason with existing parallel and efficient-reasoning methods. Parason gives the best reported AIME25 and AMC results among the listed systems, reaches the best Parason four-benchmark average accuracy of 84.7\%, and stays close to ThreadWeaver on AIME24. Across the original settings and the additional $\alpha=0.1$ runs, PA-GRPO reaches 70.6\% on AIME25, 97.5\% on AMC, and 94.6\% on Math500. These gains show that Parason improves both accuracy and average benchmark coverage over prior parallel work, even though it uses an 8B model rather than several 32B baselines.

\textbf{Parason gives a better token-latency--accuracy trade-off under constraints.} Table~\ref{tab:main-comparison} shows that Parason is most useful when the model has a small thinking budget on the longest token path. At 2048 tokens, the best Subtask-aware PA-GRPO model reaches 34.7\% AIME24 accuracy, compared with 16.8\% for SFT only. At 8192 tokens, it reaches 60.3\%, compared with 41.8\% for SFT only at the same budget. Thus, Parason does not simply spend more tokens; it moves useful work into parallel branches and keeps the longest path short.

\subsection{Insights and Findings}
\label{sec:ablations_findings}

\textbf{Trial rewards give the clearest accuracy gains, while the latency penalty shortens the critical path.} Table~\ref{tab:comparison_with_baselines} compares PA-GRPO with ThreadWeaver, which obtains 81.0\% average accuracy with a four-benchmark token latency of 14.8k. The original $\beta_{\text{trial}}=0.100$ setting gives the best average accuracy (84.7\%), while $\beta_{\text{trial}}=0.050$ gives the best AIME25 result (70.6\%). The best AIME24 result (78.2\%) is shared by the original and additional $\beta_{\text{trial}}=0.100$ settings, and the best AMC result (97.5\%) is shared by the original $\beta_{\text{trial}}=0.100$ model and the additional $\alpha=0.1$, $\beta_{\text{subtask}}=0.100$ model. Among the additional runs, $\beta_{\text{trial}}=0.025$ gives the lowest four-benchmark token latency at 12.1k. The additional sweep spans 12.1--13.4k, below ThreadWeaver's 14.8k average latency.

\begin{table}[t]
    \centering
    \footnotesize
    \setlength{\tabcolsep}{4.5pt}
    \renewcommand{\arraystretch}{1.10}
    \begin{tabular}{l c c c | c}
        \toprule
        Method & Trigger Ratio & Token Latency & Acceleration Ratio & Avg. Math Acc. \\
        \midrule
        ThreadWeaver (8B) & 83.7 & 13.3k & 1.18x & 77.6 \\
        Multiverse (32B) & 65.6 & 10.3k & 1.16x & 63.8 \\
        \midrule
        Parason-8B (w/ SFT only) & 69.3 & 14.1k & 1.27x & 78.3\\
        \addlinespace[2pt]
        \multicolumn{5}{l}{\textit{Original runs with $\alpha=0.000$}} \\
        \quad + PA-GRPO ($\beta_{\text{trial}} = 0.025$, $\alpha = 0.000$) & 77.3 & 15.2k & 1.46x & 79.9 \\
        \quad + PA-GRPO ($\beta_{\text{trial}} = 0.050$, $\alpha = 0.000$) & 84.0 & 15.8k & 1.61x & \underline{80.4} \\
        \quad + PA-GRPO ($\beta_{\text{trial}} = 0.100$, $\alpha = 0.000$) & 
        \textbf{98.8} & 15.7k & 1.71x & \textbf{80.4} \\
        \quad + PA-GRPO ($\beta_{\text{subtask}} = 0.025$, $\alpha = 0.000$) & 70.1 & 14.9k & 1.48x & 79.5 \\
        \quad + PA-GRPO ($\beta_{\text{subtask}} = 0.050$, $\alpha = 0.000$) & 79.3 & 16.6k & \textbf{1.75x} & 80.3 \\
        \quad + PA-GRPO ($\beta_{\text{subtask}} = 0.100$, $\alpha = 0.000$) & 89.6 & 15.4k & 1.73x & 80.1 \\
        \addlinespace[2pt]
        \multicolumn{5}{l}{\textit{Additional runs with $\alpha=0.100$}} \\
        \quad + PA-GRPO ($\beta_{\text{trial}} = 0.025$, $\alpha = 0.100$) & 62.3 & \textbf{13.2k} & 1.46x & 79.4 \\
        \quad + PA-GRPO ($\beta_{\text{trial}} = 0.050$, $\alpha = 0.100$) & 79.7 & 13.6k & 1.50x & 79.3 \\
        \quad + PA-GRPO ($\beta_{\text{trial}} = 0.100$, $\alpha = 0.100$) & 95.3 & 14.5k & 1.72x & 80.3 \\
        \quad + PA-GRPO ($\beta_{\text{subtask}} = 0.025$, $\alpha = 0.100$) & 53.7 & \underline{13.4k} & 1.34x & 79.0 \\
        \quad + PA-GRPO ($\beta_{\text{subtask}} = 0.050$, $\alpha = 0.100$) & 81.3 & \textbf{13.2k} & 1.60x & 78.0 \\
        \quad + PA-GRPO ($\beta_{\text{subtask}} = 0.100$, $\alpha = 0.100$) & \underline{96.5} & 14.3k & \underline{1.74x} & 78.0 \\
        \bottomrule
    \end{tabular}
    \caption{\textbf{PA-GRPO produces usable parallelism across latency-penalty settings.} For PA-GRPO, Trigger Ratio and Avg. Math Acc. are averaged over AIME24, AIME25, and Math500. \textbf{Acceleration ratio} is total tokens / longest-path tokens, and \textbf{trigger ratio} is the fraction of samples containing at least one \texttt{<Parallel>} block. Bold and underlined values mark the best and second-best PA-GRPO results among available measurements.
    }
    \label{tab:parallel_self_accel_ratio}
\end{table}

\textbf{PA-GRPO turns more parallel structure into executable acceleration.} Table~\ref{tab:parallel_self_accel_ratio} shows that the learned parallel regions are usable by the runtime. SFT alone already triggers parallel execution on 69.3\% of samples and yields 1.27$\times$ token-level acceleration. Under the original $\alpha=0.000$ setting, the Trial sweep raises the trigger ratio from 77.3\% to 98.8\% and the acceleration ratio from 1.46$\times$ to 1.71$\times$, while maintaining 79.9--80.4\% average math accuracy. Subtask incentives provide another route to high acceleration, with $\beta_{\text{subtask}}=0.050$ reaching the best acceleration ratio of 1.75$\times$ and 80.3\% average accuracy. In the additional $\alpha=0.100$ runs, acceleration rises with the parallelism coefficient: $\beta_{\text{subtask}}=0.100$ reaches 1.74$\times$ and $\beta_{\text{trial}}=0.100$ reaches 1.72$\times$. The lowest three-benchmark token latency is 13.2k, reached by $\beta_{\text{trial}}=0.025$ and $\beta_{\text{subtask}}=0.050$ after rounding.

\begin{table}[ht]
    \centering
    \setlength{\tabcolsep}{3pt}
    \renewcommand{\arraystretch}{1.15}
    \scriptsize

    \begin{tabular}{l c c c c c c c}
        \toprule
        & \multicolumn{5}{c}{Token-Level Metrics} & \multicolumn{2}{c}{Wall-Clock Metrics} \\
        \cmidrule(lr){2-6} \cmidrule(lr){7-8}
        Difficulty & \#Problems & \makecell{Generated \\ \#Tokens} & \makecell{Latency \\ \#Tokens} & \makecell{Acceleration \\ Ratio} & \makecell{Saved \\ \#Tokens} & \makecell{Parallel \\ (s)} & \makecell{Sequential \\ (s)} \\
        \midrule
        Easy   & 13 & 21.3k & 12.5k & $1.70\times$ & 8.8k  & 188.3 & 304.2 \\
        Medium & 11 & 36.8k & 21.2k & $1.74\times$ & 15.6k & 336.2 & 464.7 \\
        Hard   & 6  & 50.3k & 29.0k & $1.73\times$ & 21.3k & 487.3 & 716.3 \\
        \bottomrule
    \end{tabular}
    \caption{\textbf{Harder questions save more tokens, while parallel execution reduces wall-clock latency across all difficulty levels.} We break down AIME24 thinking tokens and measured wall-clock time by difficulty. Latency \#Tokens denotes the length of the longest token path, Acceleration Ratio is generated tokens divided by latency tokens, and Saved \#Tokens is generated tokens minus latency tokens. The experiment is executed on A800 GPUs. }
    \label{tab:token_breakdown_parallelism}
\end{table}

\textbf{Harder questions create more parallelizable work, not less acceleration.} Table~\ref{tab:token_breakdown_parallelism} breaks down the step-200 AIME24 run by difficulty. Generated tokens grow sharply with difficulty, from 21.3k on easy problems to 36.8k on medium problems and 50.3k on hard problems. The longest token path also grows, but much more slowly: 12.5k, 21.2k, and 29.0k tokens, respectively. As a result, Parason keeps a stable 1.70--1.74$\times$ token-level acceleration ratio across all three difficulty groups. This translates into measured wall-clock speedups of 1.62$\times$ on easy, 1.38$\times$ on medium, and 1.47$\times$ on hard problems. The absolute number of saved tokens increases from 8.8k on easy problems to 21.3k on hard problems, showing that harder questions expose more branch-level work that can be moved off the critical path. This supports the central motivation of Parason: difficult reasoning still requires long and diverse computation, but much of that computation does not need to remain serial.

\section{Conclusion}
\label{sec:conclusion}

In this work, we introduced \textbf{Parason}, an algorithm-system co-design framework that brings parallel processing to sequential LLM reasoning. Parason distinguishes deterministic \textbf{Subtask Parallelism} from speculative \textbf{Trial Parallelism}, and provides a stable pipeline for data curation, CFG-based structure, PA-GRPO training, and inference-engine execution. This design enables efficient parallel reasoning and improves the responsiveness of interactive agents, coding assistants, and scientific problem-solving tools. Our findings show that {Trial Parallelism} is a key mechanism for complex mathematical reasoning, although prior work has mainly focused on subtask decomposition. Across benchmarks, Parason reduces \textbf{token latency by about $1.7\times$} while \textbf{maintaining comparable accuracy}. We hope Parason raises the community's awareness of parallel reasoning and its two parallelism schemes, and inspires future work on more efficient and scalable reasoning systems. We include limitations and the impact statement in Appendix~\ref{app:limitations}, and will open-source our implementation to support reproducibility.

\begin{ack}
\end{ack}





\bibliographystyle{plainnat}
\bibliography{example_paper}

@article{dong2024xgrammar,
  title={XGrammar: Flexible and Efficient Structured Generation Engine for Large Language Models},
  author={Dong, Yixin and Ruan, Charlie F and Cai, Yaxing and Lai, Ruihang and Xu, Ziyi and Zhao, Yilong and Chen, Tianqi},
  journal={arXiv preprint arXiv:2411.15100},
  year={2024},
  url={https://arxiv.org/abs/2411.15100}
}

@article{zheng2023sglang,
  title={SGLang: Efficient Execution of Structured Language Model Programs},
  author={Zheng, Lianmin and Yin, Liangsheng and Xie, Zhiqiang and Sun, Chuyue and Huang, Jeff and Yu, Cody Hao and Cao, Shiyi and Kozyrakis, Christos and Stoica, Ion and Gonzalez, Joseph E and Barrett, Clark and Sheng, Ying},
  journal={arXiv preprint arXiv:2312.07104},
  year={2023},
  url={https://arxiv.org/abs/2312.07104}
}

@misc{deepseekai2026deepseekv4,
  title={DeepSeek-V4: Towards Highly Efficient Million-Token Context Intelligence},
  author={DeepSeek-AI},
  year={2026},
  url={https://huggingface.co/deepseek-ai/DeepSeek-V4-Pro/blob/main/DeepSeek_V4.pdf}
}

@article{zheng2025parallelr1,
  title={Parallel-r1: Towards parallel thinking via reinforcement learning},
  author={Zheng, Tong and Zhang, Hongming and Yu, Wenhao and Wang, Xiaoyang and Yang, Xinyu and Dai, Runpeng and Liu, Rui and Bao, Huiwen and Huang, Chengsong and Huang, Heng and others},
  journal={arXiv preprint arXiv:2509.07980},
  year={2025}
}

@article{openai2024o1,
  title={OpenAI o1 System Card},
  author={OpenAI and Ahmad, Lama and Askell, Amanda and Mishkin, Pamela and O'Keefe, Thomas and others},
  journal={arXiv preprint arXiv:2412.16720},
  year={2024},
  url={https://arxiv.org/abs/2412.16720}
}

@article{cobbe2021trainingverifierssolvemath,
  author       = {Karl Cobbe and
                  Vineet Kosaraju and
                  Mohammad Bavarian and
                  Mark Chen and
                  Heewoo Jun and
                  Lukasz Kaiser and
                  Matthias Plappert and
                  Jerry Tworek and
                  Jacob Hilton and
                  Reiichiro Nakano and
                  Christopher Hesse and
                  John Schulman},
  title        = {Training Verifiers to Solve Math Word Problems},
  journal      = {CoRR},
  volume       = {abs/2110.14168},
  year         = {2021},
  url          = {https://arxiv.org/abs/2110.14168},
  eprinttype    = {arXiv},
  eprint       = {2110.14168},
  bibsource    = {dblp computer science bibliography, https://dblp.org}
}

@article{wang2025survey,
  title={A Survey on Parallel Reasoning},
  author={Wang, Ziqi and Niu, Boye and Gao, Zipeng and Zheng, Zhi and Xu, Tong and Meng, Linghui and Li, Zhongli and Liu, Jing and Chen, Yilong and Zhu, Chen and others},
  journal={arXiv preprint arXiv:2510.12164},
  year={2025}
}

@inproceedings{wei2022chain,
  title={Chain-of-thought prompting elicits reasoning in large language models},
  author={Wei, Jason and Wang, Xuezhi and Schuurmans, Dale and Bosma, Maarten and Ichter, Brian and Xia, Fei and Chi, Ed and Le, Quoc V and Zhou, Denny},
  booktitle={Advances in Neural Information Processing Systems},
  volume={35},
  pages={24824--24837},
  year={2022}
}

@article{jin2025learning,
  title={Learning to keep a promise: Scaling language model decoding parallelism with learned asynchronous decoding},
  author={Jin, Tian and Cheng, Ellie Y and Ankner, Zack and Saunshi, Nikunj and Elias, Blake M and Yazdanbakhsh, Amir and Ragan-Kelley, Jonathan and Subramanian, Suvinay and Carbin, Michael},
  journal={arXiv preprint arXiv:2502.11517},
  year={2025}
}

@article{yang2025multiverse,
  title={Multiverse: Your Language Models Secretly Decide How to Parallelize and Merge Generation},
  author={Yang, Xinyu and An, Yuwei and Liu, Hongyi and Chen, Tianqi and Chen, Beidi},
  journal={arXiv preprint arXiv:2506.09991},
  year={2025}
}

@article{pan2025learning,
  title={Learning adaptive parallel reasoning with language models},
  author={Pan, Jiayi and Li, Xiuyu and Lian, Long and Snell, Charlie and Zhou, Yifei and Yala, Adam and Darrell, Trevor and Keutzer, Kurt and Suhr, Alane},
  journal={arXiv preprint arXiv:2504.15466},
  year={2025}
}

@article{sheng2024hybridflow,
  title   = {HybridFlow: A Flexible and Efficient RLHF Framework},
  author  = {Guangming Sheng and Chi Zhang and Zilingfeng Ye and Xibin Wu and Wang Zhang and Ru Zhang and Yanghua Peng and Haibin Lin and Chuan Wu},
  year    = {2024},
  journal = {arXiv preprint arXiv: 2409.19256}
}

@article{yang2025qwen3,
  title={Qwen3 technical report},
  author={Yang, An and Li, Anfeng and Yang, Baosong and Zhang, Beichen and Hui, Binyuan and Zheng, Bo and Yu, Bowen and Gao, Chang and Huang, Chengen and Lv, Chenxu and others},
  journal={arXiv preprint arXiv:2505.09388},
  year={2025}
}

@article{yao2023tree,
  title={Tree of thoughts: Deliberate problem solving with large language models},
  author={Yao, Shunyu and Yu, Dian and Zhao, Jeffrey and Shafran, Izhak and Griffiths, Tom and Cao, Yuan and Narasimhan, Karthik},
  journal={Advances in neural information processing systems},
  volume={36},
  pages={11809--11822},
  year={2023}
}

@article{wang2022self,
  title={Self-consistency improves chain of thought reasoning in language models},
  author={Wang, Xuezhi and Wei, Jason and Schuurmans, Dale and Le, Quoc and Chi, Ed and Narang, Sharan and Chowdhery, Aakanksha and Zhou, Denny},
  journal={arXiv preprint arXiv:2203.11171},
  year={2022}
}

@article{fu2025deep,
  title={Deep Think with Confidence},
  author={Fu, Yichao and Wang, Xuewei and Tian, Yuandong and Zhao, Jiawei},
  journal={arXiv preprint arXiv:2508.15260},
  year={2025}
}

@misc{imojourney,
      title={Towards Robust Mathematical Reasoning}, 
      author={Thang Luong and Dawsen Hwang and Hoang H. Nguyen and Golnaz Ghiasi and Yuri Chervonyi and Insuk Seo and Junsu Kim and Garrett Bingham and Jonathan Lee and Swaroop Mishra and Alex Zhai and Clara Huiyi Hu and Henryk Michalewski and Jimin Kim and Jeonghyun Ahn and Junhwi Bae and Xingyou Song and Trieu H. Trinh and Quoc V. Le and Junehyuk Jung},
      year={2025},
      eprint={2511.01846},
      archivePrefix={arXiv},
      primaryClass={cs.CL},
      url={https://arxiv.org/abs/2511.01846}, 
}

@misc{pareco,
      title={PaCoRe: Learning to Scale Test-Time Compute with Parallel Coordinated Reasoning}, 
      author={Jingcheng Hu and Yinmin Zhang and Shijie Shang and Xiaobo Yang and Yue Peng and Zhewei Huang and Hebin Zhou and Xin Wu and Jie Cheng and Fanqi Wan and Xiangwen Kong and Chengyuan Yao and Kaiwen Yan and Ailin Huang and Hongyu Zhou and Qi Han and Zheng Ge and Daxin Jiang and Xiangyu Zhang and Heung-Yeung Shum},
      year={2026},
      eprint={2601.05593},
      archivePrefix={arXiv},
      primaryClass={cs.LG},
      url={https://arxiv.org/abs/2601.05593}, 
}

@article{imo24,
  title = {Olympiad-level formal mathematical reasoning with reinforcement learning},
  ISSN = {1476-4687},
  url = {http://dx.doi.org/10.1038/s41586-025-09833-y},
  DOI = {10.1038/s41586-025-09833-y},
  journal = {Nature},
  publisher = {Springer Science and Business Media LLC},
  author = {Hubert,  Thomas and Mehta,  Rishi and Sartran,  Laurent and Horváth,  Miklós Z. and Žužić,  Goran and Wieser,  Eric and Huang,  Aja and Schrittwieser,  Julian and Schroecker,  Yannick and Masoom,  Hussain and Bertolli,  Ottavia and Zahavy,  Tom and Mandhane,  Amol and Yung,  Jessica and Beloshapka,  Iuliya and Ibarz,  Borja and Veeriah,  Vivek and Yu,  Lei and Nash,  Oliver and Lezeau,  Paul and Mercuri,  Salvatore and S\"{o}nne,  Calle and Mehta,  Bhavik and Davies,  Alex and Zheng,  Daniel and Pedregosa,  Fabian and Li,  Yin and von Glehn,  Ingrid and Rowland,  Mark and Albanie,  Samuel and Velingker,  Ameya and Schmitt,  Simon and Lockhart,  Edward and Hughes,  Edward and Michalewski,  Henryk and Sonnerat,  Nicolas and Hassabis,  Demis and Kohli,  Pushmeet and Silver,  David},
  year = {2025},
  month = nov 
}

@misc{openmath,
      title={AIMO-2 Winning Solution: Building State-of-the-Art Mathematical Reasoning Models with OpenMathReasoning dataset}, 
      author={Ivan Moshkov and Darragh Hanley and Ivan Sorokin and Shubham Toshniwal and Christof Henkel and Benedikt Schifferer and Wei Du and Igor Gitman},
      year={2025},
      eprint={2504.16891},
      archivePrefix={arXiv},
      primaryClass={cs.AI},
      url={https://arxiv.org/abs/2504.16891}, 
}

@misc{hle25,
      title={Humanity's Last Exam}, 
      author={Long Phan and Alice Gatti and Ziwen Han and Nathaniel Li and Josephina Hu and Hugh Zhang and Chen Bo Calvin Zhang and Mohamed Shaaban and John Ling and Sean Shi and Michael Choi and Anish Agrawal and Arnav Chopra and Adam Khoja and Ryan Kim and Richard Ren and Jason Hausenloy and Oliver Zhang and Mantas Mazeika and Dmitry Dodonov and Tung Nguyen and Jaeho Lee and Daron Anderson and Mikhail Doroshenko and Alun Cennyth Stokes and Mobeen Mahmood and Oleksandr Pokutnyi and Oleg Iskra and Jessica P. Wang and John-Clark Levin and Mstyslav Kazakov and Fiona Feng and Steven Y. Feng and Haoran Zhao and Michael Yu and Varun Gangal and Chelsea Zou and Zihan Wang and Serguei Popov and Robert Gerbicz and Geoff Galgon and Johannes Schmitt and Will Yeadon and Yongki Lee and Scott Sauers and Alvaro Sanchez and Fabian Giska and Marc Roth and Søren Riis and Saiteja Utpala and Noah Burns and Gashaw M. Goshu and Mohinder Maheshbhai Naiya and Chidozie Agu and Zachary Giboney and Antrell Cheatom and Francesco Fournier-Facio and Sarah-Jane Crowson and Lennart Finke and Zerui Cheng and Jennifer Zampese and Ryan G. Hoerr and Mark Nandor and Hyunwoo Park and Tim Gehrunger and Jiaqi Cai and Ben McCarty and Alexis C Garretson and Edwin Taylor and Damien Sileo and Qiuyu Ren and Usman Qazi and Lianghui Li and Jungbae Nam and John B. Wydallis and Pavel Arkhipov and Jack Wei Lun Shi and Aras Bacho and Chris G. Willcocks and Hangrui Cao and Sumeet Motwani and Emily de Oliveira Santos and Johannes Veith and Edward Vendrow and Doru Cojoc and Kengo Zenitani and Joshua Robinson and Longke Tang and Yuqi Li and Joshua Vendrow and Natanael Wildner Fraga and Vladyslav Kuchkin and Andrey Pupasov Maksimov and Pierre Marion and Denis Efremov and Jayson Lynch and Kaiqu Liang and Aleksandar Mikov and Andrew Gritsevskiy and Julien Guillod and Gözdenur Demir and Dakotah Martinez and Ben Pageler and Kevin Zhou and Saeed Soori and Ori Press and Henry Tang and Paolo Rissone and Sean R. Green and Lina Brüssel and Moon Twayana and Aymeric Dieuleveut and Joseph Marvin Imperial and Ameya Prabhu and Jinzhou Yang and Nick Crispino and Arun Rao and Dimitri Zvonkine and Gabriel Loiseau and Mikhail Kalinin and Marco Lukas and Ciprian Manolescu and Nate Stambaugh and Subrata Mishra and Tad Hogg and Carlo Bosio and Brian P Coppola and Julian Salazar and Jaehyeok Jin and Rafael Sayous and Stefan Ivanov and Philippe Schwaller and Shaipranesh Senthilkuma and Andres M Bran and Andres Algaba and Kelsey Van den Houte and Lynn Van Der Sypt and Brecht Verbeken and David Noever and Alexei Kopylov and Benjamin Myklebust and Bikun Li and Lisa Schut and Evgenii Zheltonozhskii and Qiaochu Yuan and Derek Lim and Richard Stanley and Tong Yang and John Maar and Julian Wykowski and Martí Oller and Anmol Sahu and Cesare Giulio Ardito and Yuzheng Hu and Ariel Ghislain Kemogne Kamdoum and Alvin Jin and Tobias Garcia Vilchis and Yuexuan Zu and Martin Lackner and James Koppel and Gongbo Sun and Daniil S. Antonenko and Steffi Chern and Bingchen Zhao and Pierrot Arsene and Joseph M Cavanagh and Daofeng Li and Jiawei Shen and Donato Crisostomi and Wenjin Zhang and Ali Dehghan and Sergey Ivanov and David Perrella and Nurdin Kaparov and Allen Zang and Ilia Sucholutsky and Arina Kharlamova and Daniil Orel and Vladislav Poritski and Shalev Ben-David and Zachary Berger and Parker Whitfill and Michael Foster and Daniel Munro and Linh Ho and Shankar Sivarajan and Dan Bar Hava and Aleksey Kuchkin and David Holmes and Alexandra Rodriguez-Romero and Frank Sommerhage and Anji Zhang and Richard Moat and Keith Schneider and Zakayo Kazibwe and Don Clarke and Dae Hyun Kim and Felipe Meneguitti Dias and Sara Fish and Veit Elser and Tobias Kreiman and Victor Efren Guadarrama Vilchis and Immo Klose and Ujjwala Anantheswaran and Adam Zweiger and Kaivalya Rawal and Jeffery Li and Jeremy Nguyen and Nicolas Daans and Haline Heidinger and Maksim Radionov and Václav Rozhoň and Vincent Ginis and Christian Stump and Niv Cohen and Rafał Poświata and Josef Tkadlec and Alan Goldfarb and Chenguang Wang and Piotr Padlewski and Stanislaw Barzowski and Kyle Montgomery and Ryan Stendall and Jamie Tucker-Foltz and Jack Stade and T. Ryan Rogers and Tom Goertzen and Declan Grabb and Abhishek Shukla and Alan Givré and John Arnold Ambay and Archan Sen and Muhammad Fayez Aziz and Mark H Inlow and Hao He and Ling Zhang and Younesse Kaddar and Ivar Ängquist and Yanxu Chen and Harrison K Wang and Kalyan Ramakrishnan and Elliott Thornley and Antonio Terpin and Hailey Schoelkopf and Eric Zheng and Avishy Carmi and Ethan D. L. Brown and Kelin Zhu and Max Bartolo and Richard Wheeler and Martin Stehberger and Peter Bradshaw and JP Heimonen and Kaustubh Sridhar and Ido Akov and Jennifer Sandlin and Yury Makarychev and Joanna Tam and Hieu Hoang and David M. Cunningham and Vladimir Goryachev and Demosthenes Patramanis and Michael Krause and Andrew Redenti and David Aldous and Jesyin Lai and Shannon Coleman and Jiangnan Xu and Sangwon Lee and Ilias Magoulas and Sandy Zhao and Ning Tang and Michael K. Cohen and Orr Paradise and Jan Hendrik Kirchner and Maksym Ovchynnikov and Jason O. Matos and Adithya Shenoy and Michael Wang and Yuzhou Nie and Anna Sztyber-Betley and Paolo Faraboschi and Robin Riblet and Jonathan Crozier and Shiv Halasyamani and Shreyas Verma and Prashant Joshi and Eli Meril and Ziqiao Ma and Jérémy Andréoletti and Raghav Singhal and Jacob Platnick and Volodymyr Nevirkovets and Luke Basler and Alexander Ivanov and Seri Khoury and Nils Gustafsson and Marco Piccardo and Hamid Mostaghimi and Qijia Chen and Virendra Singh and Tran Quoc Khánh and Paul Rosu and Hannah Szlyk and Zachary Brown and Himanshu Narayan and Aline Menezes and Jonathan Roberts and William Alley and Kunyang Sun and Arkil Patel and Max Lamparth and Anka Reuel and Linwei Xin and Hanmeng Xu and Jacob Loader and Freddie Martin and Zixuan Wang and Andrea Achilleos and Thomas Preu and Tomek Korbak and Ida Bosio and Fereshteh Kazemi and Ziye Chen and Biró Bálint and Eve J. Y. Lo and Jiaqi Wang and Maria Inês S. Nunes and Jeremiah Milbauer and M Saiful Bari and Zihao Wang and Behzad Ansarinejad and Yewen Sun and Stephane Durand and Hossam Elgnainy and Guillaume Douville and Daniel Tordera and George Balabanian and Hew Wolff and Lynna Kvistad and Hsiaoyun Milliron and Ahmad Sakor and Murat Eron and Andrew Favre D. O. and Shailesh Shah and Xiaoxiang Zhou and Firuz Kamalov and Sherwin Abdoli and Tim Santens and Shaul Barkan and Allison Tee and Robin Zhang and Alessandro Tomasiello and G. Bruno De Luca and Shi-Zhuo Looi and Vinh-Kha Le and Noam Kolt and Jiayi Pan and Emma Rodman and Jacob Drori and Carl J Fossum and Niklas Muennighoff and Milind Jagota and Ronak Pradeep and Honglu Fan and Jonathan Eicher and Michael Chen and Kushal Thaman and William Merrill and Moritz Firsching and Carter Harris and Stefan Ciobâcă and Jason Gross and Rohan Pandey and Ilya Gusev and Adam Jones and Shashank Agnihotri and Pavel Zhelnov and Mohammadreza Mofayezi and Alexander Piperski and David K. Zhang and Kostiantyn Dobarskyi and Roman Leventov and Ignat Soroko and Joshua Duersch and Vage Taamazyan and Andrew Ho and Wenjie Ma and William Held and Ruicheng Xian and Armel Randy Zebaze and Mohanad Mohamed and Julian Noah Leser and Michelle X Yuan and Laila Yacar and Johannes Lengler and Katarzyna Olszewska and Claudio Di Fratta and Edson Oliveira and Joseph W. Jackson and Andy Zou and Muthu Chidambaram and Timothy Manik and Hector Haffenden and Dashiell Stander and Ali Dasouqi and Alexander Shen and Bita Golshani and David Stap and Egor Kretov and Mikalai Uzhou and Alina Borisovna Zhidkovskaya and Nick Winter and Miguel Orbegozo Rodriguez and Robert Lauff and Dustin Wehr and Colin Tang and Zaki Hossain and Shaun Phillips and Fortuna Samuele and Fredrik Ekström and Angela Hammon and Oam Patel and Faraz Farhidi and George Medley and Forough Mohammadzadeh and Madellene Peñaflor and Haile Kassahun and Alena Friedrich and Rayner Hernandez Perez and Daniel Pyda and Taom Sakal and Omkar Dhamane and Ali Khajegili Mirabadi and Eric Hallman and Kenchi Okutsu and Mike Battaglia and Mohammad Maghsoudimehrabani and Alon Amit and Dave Hulbert and Roberto Pereira and Simon Weber and Handoko and Anton Peristyy and Stephen Malina and Mustafa Mehkary and Rami Aly and Frank Reidegeld and Anna-Katharina Dick and Cary Friday and Mukhwinder Singh and Hassan Shapourian and Wanyoung Kim and Mariana Costa and Hubeyb Gurdogan and Harsh Kumar and Chiara Ceconello and Chao Zhuang and Haon Park and Micah Carroll and Andrew R. Tawfeek and Stefan Steinerberger and Daattavya Aggarwal and Michael Kirchhof and Linjie Dai and Evan Kim and Johan Ferret and Jainam Shah and Yuzhou Wang and Minghao Yan and Krzysztof Burdzy and Lixin Zhang and Antonio Franca and Diana T. Pham and Kang Yong Loh and Joshua Robinson and Abram Jackson and Paolo Giordano and Philipp Petersen and Adrian Cosma and Jesus Colino and Colin White and Jacob Votava and Vladimir Vinnikov and Ethan Delaney and Petr Spelda and Vit Stritecky and Syed M. Shahid and Jean-Christophe Mourrat and Lavr Vetoshkin and Koen Sponselee and Renas Bacho and Zheng-Xin Yong and Florencia de la Rosa and Nathan Cho and Xiuyu Li and Guillaume Malod and Orion Weller and Guglielmo Albani and Leon Lang and Julien Laurendeau and Dmitry Kazakov and Fatimah Adesanya and Julien Portier and Lawrence Hollom and Victor Souza and Yuchen Anna Zhou and Julien Degorre and Yiğit Yalın and Gbenga Daniel Obikoya and Rai and Filippo Bigi and M. C. Boscá and Oleg Shumar and Kaniuar Bacho and Gabriel Recchia and Mara Popescu and Nikita Shulga and Ngefor Mildred Tanwie and Thomas C. H. Lux and Ben Rank and Colin Ni and Matthew Brooks and Alesia Yakimchyk and Huanxu and Liu and Stefano Cavalleri and Olle Häggström and Emil Verkama and Joshua Newbould and Hans Gundlach and Leonor Brito-Santana and Brian Amaro and Vivek Vajipey and Rynaa Grover and Ting Wang and Yosi Kratish and Wen-Ding Li and Sivakanth Gopi and Andrea Caciolai and Christian Schroeder de Witt and Pablo Hernández-Cámara and Emanuele Rodolà and Jules Robins and Dominic Williamson and Vincent Cheng and Brad Raynor and Hao Qi and Ben Segev and Jingxuan Fan and Sarah Martinson and Erik Y. Wang and Kaylie Hausknecht and Michael P. Brenner and Mao Mao and Christoph Demian and Peyman Kassani and Xinyu Zhang and David Avagian and Eshawn Jessica Scipio and Alon Ragoler and Justin Tan and Blake Sims and Rebeka Plecnik and Aaron Kirtland and Omer Faruk Bodur and D. P. Shinde and Yan Carlos Leyva Labrador and Zahra Adoul and Mohamed Zekry and Ali Karakoc and Tania C. B. Santos and Samir Shamseldeen and Loukmane Karim and Anna Liakhovitskaia and Nate Resman and Nicholas Farina and Juan Carlos Gonzalez and Gabe Maayan and Earth Anderson and Rodrigo De Oliveira Pena and Elizabeth Kelley and Hodjat Mariji and Rasoul Pouriamanesh and Wentao Wu and Ross Finocchio and Ismail Alarab and Joshua Cole and Danyelle Ferreira and Bryan Johnson and Mohammad Safdari and Liangti Dai and Siriphan Arthornthurasuk and Isaac C. McAlister and Alejandro José Moyano and Alexey Pronin and Jing Fan and Angel Ramirez-Trinidad and Yana Malysheva and Daphiny Pottmaier and Omid Taheri and Stanley Stepanic and Samuel Perry and Luke Askew and Raúl Adrián Huerta Rodríguez and Ali M. R. Minissi and Ricardo Lorena and Krishnamurthy Iyer and Arshad Anil Fasiludeen and Ronald Clark and Josh Ducey and Matheus Piza and Maja Somrak and Eric Vergo and Juehang Qin and Benjámin Borbás and Eric Chu and Jack Lindsey and Antoine Jallon and I. M. J. McInnis and Evan Chen and Avi Semler and Luk Gloor and Tej Shah and Marc Carauleanu and Pascal Lauer and Tran Đuc Huy and Hossein Shahrtash and Emilien Duc and Lukas Lewark and Assaf Brown and Samuel Albanie and Brian Weber and Warren S. Vaz and Pierre Clavier and Yiyang Fan and Gabriel Poesia Reis e Silva and Long and Lian and Marcus Abramovitch and Xi Jiang and Sandra Mendoza and Murat Islam and Juan Gonzalez and Vasilios Mavroudis and Justin Xu and Pawan Kumar and Laxman Prasad Goswami and Daniel Bugas and Nasser Heydari and Ferenc Jeanplong and Thorben Jansen and Antonella Pinto and Archimedes Apronti and Abdallah Galal and Ng Ze-An and Ankit Singh and Tong Jiang and Joan of Arc Xavier and Kanu Priya Agarwal and Mohammed Berkani and Gang Zhang and Zhehang Du and Benedito Alves de Oliveira Junior and Dmitry Malishev and Nicolas Remy and Taylor D. Hartman and Tim Tarver and Stephen Mensah and Gautier Abou Loume and Wiktor Morak and Farzad Habibi and Sarah Hoback and Will Cai and Javier Gimenez and Roselynn Grace Montecillo and Jakub Łucki and Russell Campbell and Asankhaya Sharma and Khalida Meer and Shreen Gul and Daniel Espinosa Gonzalez and Xavier Alapont and Alex Hoover and Gunjan Chhablani and Freddie Vargus and Arunim Agarwal and Yibo Jiang and Deepakkumar Patil and David Outevsky and Kevin Joseph Scaria and Rajat Maheshwari and Abdelkader Dendane and Priti Shukla and Ashley Cartwright and Sergei Bogdanov and Niels Mündler and Sören Möller and Luca Arnaboldi and Kunvar Thaman and Muhammad Rehan Siddiqi and Prajvi Saxena and Himanshu Gupta and Tony Fruhauff and Glen Sherman and Mátyás Vincze and Siranut Usawasutsakorn and Dylan Ler and Anil Radhakrishnan and Innocent Enyekwe and Sk Md Salauddin and Jiang Muzhen and Aleksandr Maksapetyan and Vivien Rossbach and Chris Harjadi and Mohsen Bahaloohoreh and Claire Sparrow and Jasdeep Sidhu and Sam Ali and Song Bian and John Lai and Eric Singer and Justine Leon Uro and Greg Bateman and Mohamed Sayed and Ahmed Menshawy and Darling Duclosel and Dario Bezzi and Yashaswini Jain and Ashley Aaron and Murat Tiryakioglu and Sheeshram Siddh and Keith Krenek and Imad Ali Shah and Jun Jin and Scott Creighton and Denis Peskoff and Zienab EL-Wasif and Ragavendran P V and Michael Richmond and Joseph McGowan and Tejal Patwardhan and Hao-Yu Sun and Ting Sun and Nikola Zubić and Samuele Sala and Stephen Ebert and Jean Kaddour and Manuel Schottdorf and Dianzhuo Wang and Gerol Petruzella and Alex Meiburg and Tilen Medved and Ali ElSheikh and S Ashwin Hebbar and Lorenzo Vaquero and Xianjun Yang and Jason Poulos and Vilém Zouhar and Sergey Bogdanik and Mingfang Zhang and Jorge Sanz-Ros and David Anugraha and Yinwei Dai and Anh N. Nhu and Xue Wang and Ali Anil Demircali and Zhibai Jia and Yuyin Zhou and Juncheng Wu and Mike He and Nitin Chandok and Aarush Sinha and Gaoxiang Luo and Long Le and Mickaël Noyé and Michał Perełkiewicz and Ioannis Pantidis and Tianbo Qi and Soham Sachin Purohit and Letitia Parcalabescu and Thai-Hoa Nguyen and Genta Indra Winata and Edoardo M. Ponti and Hanchen Li and Kaustubh Dhole and Jongee Park and Dario Abbondanza and Yuanli Wang and Anupam Nayak and Diogo M. Caetano and Antonio A. W. L. Wong and Maria del Rio-Chanona and Dániel Kondor and Pieter Francois and Ed Chalstrey and Jakob Zsambok and Dan Hoyer and Jenny Reddish and Jakob Hauser and Francisco-Javier Rodrigo-Ginés and Suchandra Datta and Maxwell Shepherd and Thom Kamphuis and Qizheng Zhang and Hyunjun Kim and Ruiji Sun and Jianzhu Yao and Franck Dernoncourt and Satyapriya Krishna and Sina Rismanchian and Bonan Pu and Francesco Pinto and Yingheng Wang and Kumar Shridhar and Kalon J. Overholt and Glib Briia and Hieu Nguyen and David and Soler Bartomeu and Tony CY Pang and Adam Wecker and Yifan Xiong and Fanfei Li and Lukas S. Huber and Joshua Jaeger and Romano De Maddalena and Xing Han Lù and Yuhui Zhang and Claas Beger and Patrick Tser Jern Kon and Sean Li and Vivek Sanker and Ming Yin and Yihao Liang and Xinlu Zhang and Ankit Agrawal and Li S. Yifei and Zechen Zhang and Mu Cai and Yasin Sonmez and Costin Cozianu and Changhao Li and Alex Slen and Shoubin Yu and Hyun Kyu Park and Gabriele Sarti and Marcin Briański and Alessandro Stolfo and Truong An Nguyen and Mike Zhang and Yotam Perlitz and Jose Hernandez-Orallo and Runjia Li and Amin Shabani and Felix Juefei-Xu and Shikhar Dhingra and Orr Zohar and My Chiffon Nguyen and Alexander Pondaven and Abdurrahim Yilmaz and Xuandong Zhao and Chuanyang Jin and Muyan Jiang and Stefan Todoran and Xinyao Han and Jules Kreuer and Brian Rabern and Anna Plassart and Martino Maggetti and Luther Yap and Robert Geirhos and Jonathon Kean and Dingsu Wang and Sina Mollaei and Chenkai Sun and Yifan Yin and Shiqi Wang and Rui Li and Yaowen Chang and Anjiang Wei and Alice Bizeul and Xiaohan Wang and Alexandre Oliveira Arrais and Kushin Mukherjee and Jorge Chamorro-Padial and Jiachen Liu and Xingyu Qu and Junyi Guan and Adam Bouyamourn and Shuyu Wu and Martyna Plomecka and Junda Chen and Mengze Tang and Jiaqi Deng and Shreyas Subramanian and Haocheng Xi and Haoxuan Chen and Weizhi Zhang and Yinuo Ren and Haoqin Tu and Sejong Kim and Yushun Chen and Sara Vera Marjanović and Junwoo Ha and Grzegorz Luczyna and Jeff J. Ma and Zewen Shen and Dawn Song and Cedegao E. Zhang and Zhun Wang and Gaël Gendron and Yunze Xiao and Leo Smucker and Erica Weng and Kwok Hao Lee and Zhe Ye and Stefano Ermon and Ignacio D. Lopez-Miguel and Theo Knights and Anthony Gitter and Namkyu Park and Boyi Wei and Hongzheng Chen and Kunal Pai and Ahmed Elkhanany and Han Lin and Philipp D. Siedler and Jichao Fang and Ritwik Mishra and Károly Zsolnai-Fehér and Xilin Jiang and Shadab Khan and Jun Yuan and Rishab Kumar Jain and Xi Lin and Mike Peterson and Zhe Wang and Aditya Malusare and Maosen Tang and Isha Gupta and Ivan Fosin and Timothy Kang and Barbara Dworakowska and Kazuki Matsumoto and Guangyao Zheng and Gerben Sewuster and Jorge Pretel Villanueva and Ivan Rannev and Igor Chernyavsky and Jiale Chen and Deepayan Banik and Ben Racz and Wenchao Dong and Jianxin Wang and Laila Bashmal and Duarte V. Gonçalves and Wei Hu and Kaushik Bar and Ondrej Bohdal and Atharv Singh Patlan and Shehzaad Dhuliawala and Caroline Geirhos and Julien Wist and Yuval Kansal and Bingsen Chen and Kutay Tire and Atak Talay Yücel and Brandon Christof and Veerupaksh Singla and Zijian Song and Sanxing Chen and Jiaxin Ge and Kaustubh Ponkshe and Isaac Park and Tianneng Shi and Martin Q. Ma and Joshua Mak and Sherwin Lai and Antoine Moulin and Zhuo Cheng and Zhanda Zhu and Ziyi Zhang and Vaidehi Patil and Ketan Jha and Qiutong Men and Jiaxuan Wu and Tianchi Zhang and Bruno Hebling Vieira and Alham Fikri Aji and Jae-Won Chung and Mohammed Mahfoud and Ha Thi Hoang and Marc Sperzel and Wei Hao and Kristof Meding and Sihan Xu and Vassilis Kostakos and Davide Manini and Yueying Liu and Christopher Toukmaji and Jay Paek and Eunmi Yu and Arif Engin Demircali and Zhiyi Sun and Ivan Dewerpe and Hongsen Qin and Roman Pflugfelder and James Bailey and Johnathan Morris and Ville Heilala and Sybille Rosset and Zishun Yu and Peter E. Chen and Woongyeong Yeo and Eeshaan Jain and Ryan Yang and Sreekar Chigurupati and Julia Chernyavsky and Sai Prajwal Reddy and Subhashini Venugopalan and Hunar Batra and Core Francisco Park and Hieu Tran and Guilherme Maximiano and Genghan Zhang and Yizhuo Liang and Hu Shiyu and Rongwu Xu and Rui Pan and Siddharth Suresh and Ziqi Liu and Samaksh Gulati and Songyang Zhang and Peter Turchin and Christopher W. Bartlett and Christopher R. Scotese and Phuong M. Cao and Ben Wu and Jacek Karwowski and Davide Scaramuzza and Aakaash Nattanmai and Gordon McKellips and Anish Cheraku and Asim Suhail and Ethan Luo and Marvin Deng and Jason Luo and Ashley Zhang and Kavin Jindel and Jay Paek and Kasper Halevy and Allen Baranov and Michael Liu and Advaith Avadhanam and David Zhang and Vincent Cheng and Brad Ma and Evan Fu and Liam Do and Joshua Lass and Hubert Yang and Surya Sunkari and Vishruth Bharath and Violet Ai and James Leung and Rishit Agrawal and Alan Zhou and Kevin Chen and Tejas Kalpathi and Ziqi Xu and Gavin Wang and Tyler Xiao and Erik Maung and Sam Lee and Ryan Yang and Roy Yue and Ben Zhao and Julia Yoon and Sunny Sun and Aryan Singh and Ethan Luo and Clark Peng and Tyler Osbey and Taozhi Wang and Daryl Echeazu and Hubert Yang and Timothy Wu and Spandan Patel and Vidhi Kulkarni and Vijaykaarti Sundarapandiyan and Ashley Zhang and Andrew Le and Zafir Nasim and Srikar Yalam and Ritesh Kasamsetty and Soham Samal and Hubert Yang and David Sun and Nihar Shah and Abhijeet Saha and Alex Zhang and Leon Nguyen and Laasya Nagumalli and Kaixin Wang and Alan Zhou and Aidan Wu and Jason Luo and Anwith Telluri and Summer Yue and Alexandr Wang and Dan Hendrycks},
      year={2025},
      eprint={2501.14249},
      archivePrefix={arXiv},
      primaryClass={cs.LG},
      url={https://arxiv.org/abs/2501.14249}, 
}

@misc{Polaris2025,
    title = {POLARIS: A Post-Training Recipe for Scaling Reinforcement Learning on Advanced Reasoning Models},
    url = {https://hkunlp.github.io/blog/2025/Polaris},
    author = {An, Chenxin and Xie, Zhihui and Li, Xiaonan and Li, Lei and Zhang, Jun and Gong, Shansan and Zhong, Ming and Xu, Jingjing and Qiu, Xipeng and Wang, Mingxuan and Kong, Lingpeng},
    year = {2025}
}

@misc{deepseekai2025deepseekr1incentivizingreasoningcapability,
      title={DeepSeek-R1: Incentivizing Reasoning Capability in LLMs via Reinforcement Learning}, 
      author={DeepSeek-AI and Daya Guo and Dejian Yang and Haowei Zhang and Junxiao Song and Ruoyu Zhang and Runxin Xu and Qihao Zhu and Shirong Ma and Peiyi Wang and Xiao Bi and Xiaokang Zhang and Xingkai Yu and Yu Wu and Z. F. Wu and Zhibin Gou and Zhihong Shao and Zhuoshu Li and Ziyi Gao and Aixin Liu and Bing Xue and Bingxuan Wang and Bochao Wu and Bei Feng and Chengda Lu and Chenggang Zhao and Chengqi Deng and Chenyu Zhang and Chong Ruan and Damai Dai and Deli Chen and Dongjie Ji and Erhang Li and Fangyun Lin and Fucong Dai and Fuli Luo and Guangbo Hao and Guanting Chen and Guowei Li and H. Zhang and Han Bao and Hanwei Xu and Haocheng Wang and Honghui Ding and Huajian Xin and Huazuo Gao and Hui Qu and Hui Li and Jianzhong Guo and Jiashi Li and Jiawei Wang and Jingchang Chen and Jingyang Yuan and Junjie Qiu and Junlong Li and J. L. Cai and Jiaqi Ni and Jian Liang and Jin Chen and Kai Dong and Kai Hu and Kaige Gao and Kang Guan and Kexin Huang and Kuai Yu and Lean Wang and Lecong Zhang and Liang Zhao and Litong Wang and Liyue Zhang and Lei Xu and Leyi Xia and Mingchuan Zhang and Minghua Zhang and Minghui Tang and Meng Li and Miaojun Wang and Mingming Li and Ning Tian and Panpan Huang and Peng Zhang and Qiancheng Wang and Qinyu Chen and Qiushi Du and Ruiqi Ge and Ruisong Zhang and Ruizhe Pan and Runji Wang and R. J. Chen and R. L. Jin and Ruyi Chen and Shanghao Lu and Shangyan Zhou and Shanhuang Chen and Shengfeng Ye and Shiyu Wang and Shuiping Yu and Shunfeng Zhou and Shuting Pan and S. S. Li and Shuang Zhou and Shaoqing Wu and Shengfeng Ye and Tao Yun and Tian Pei and Tianyu Sun and T. Wang and Wangding Zeng and Wanjia Zhao and Wen Liu and Wenfeng Liang and Wenjun Gao and Wenqin Yu and Wentao Zhang and W. L. Xiao and Wei An and Xiaodong Liu and Xiaohan Wang and Xiaokang Chen and Xiaotao Nie and Xin Cheng and Xin Liu and Xin Xie and Xingchao Liu and Xinyu Yang and Xinyuan Li and Xuecheng Su and Xuheng Lin and X. Q. Li and Xiangyue Jin and Xiaojin Shen and Xiaosha Chen and Xiaowen Sun and Xiaoxiang Wang and Xinnan Song and Xinyi Zhou and Xianzu Wang and Xinxia Shan and Y. K. Li and Y. Q. Wang and Y. X. Wei and Yang Zhang and Yanhong Xu and Yao Li and Yao Zhao and Yaofeng Sun and Yaohui Wang and Yi Yu and Yichao Zhang and Yifan Shi and Yiliang Xiong and Ying He and Yishi Piao and Yisong Wang and Yixuan Tan and Yiyang Ma and Yiyuan Liu and Yongqiang Guo and Yuan Ou and Yuduan Wang and Yue Gong and Yuheng Zou and Yujia He and Yunfan Xiong and Yuxiang Luo and Yuxiang You and Yuxuan Liu and Yuyang Zhou and Y. X. Zhu and Yanhong Xu and Yanping Huang and Yaohui Li and Yi Zheng and Yuchen Zhu and Yunxian Ma and Ying Tang and Yukun Zha and Yuting Yan and Z. Z. Ren and Zehui Ren and Zhangli Sha and Zhe Fu and Zhean Xu and Zhenda Xie and Zhengyan Zhang and Zhewen Hao and Zhicheng Ma and Zhigang Yan and Zhiyu Wu and Zihui Gu and Zijia Zhu and Zijun Liu and Zilin Li and Ziwei Xie and Ziyang Song and Zizheng Pan and Zhen Huang and Zhipeng Xu and Zhongyu Zhang and Zhen Zhang},
      year={2025},
      eprint={2501.12948},
      archivePrefix={arXiv},
      primaryClass={cs.CL},
      url={https://arxiv.org/abs/2501.12948}, 
}

@misc{google-gemini-2.5,
  author       = {Google},
  title        = {Gemini 2.5: Our most intelligent AI model},
  month        = Mar,
  year         = {2025},
  url          = {https://blog.google/technology/google-deepmind/gemini-model-thinking-updates-march-2025/\#gemini-2-5-thinking}
}

@article{ning2023skeleton,
  title={Skeleton-of-thought: Large language models can do parallel decoding},
  author={Ning, Xuefei and Lin, Zinan and Zhou, Zixuan and Wang, Zifu and Yang, Huazhong and Wang, Yu},
  journal={Proceedings ENLSP-III},
  year={2023}
}

@misc{aime2024,
  author       = {{Mathematical Association of America}},
  title        = {{American Invitational Mathematics Examination 2024}},
  year         = {2024},
  url          = {https://artofproblemsolving.com/wiki/index.php/American_Invitational_Mathematics_Examination},
  note         = {Accessed: 2025-05-14}
}

@misc{aime2025,
  author       = {{Mathematical Association of America}},
  title        = {{American Invitational Mathematics Examination 2025}},
  year         = {2025},
  url          = {https://artofproblemsolving.com/wiki/index.php/American_Invitational_Mathematics_Examination},
  note         = {Accessed: 2025-05-14}
}

@misc{amc2023,
  author       = {{Mathematical Association of America}},
  title        = {{American Mathematics Competitions 2023}},
  year         = {2023},
  url          = {https://maa.org/student-programs/amc/},
  note         = {Accessed: 2026-05-07}
}

@misc{lewkowycz2022minerva,
  title        = {Solving Quantitative Reasoning Problems with Language Models},
  author       = {Lewkowycz, Aitor and Andreassen, Anders and Dohan, David and Dyer, Ethan and Michalewski, Henryk and Ramasesh, Vinay and Slone, Ambrose and Anil, Cem and Schlag, Imanol and Gutman-Solo, Theo and Wu, Yuhuai and Neyshabur, Behnam and Gur-Ari, Guy and Misra, Vedant},
  year         = {2022},
  eprint       = {2206.14858},
  archivePrefix = {arXiv},
  primaryClass = {cs.CL},
  url          = {https://arxiv.org/abs/2206.14858}
}

@misc{math500,
      title={Measuring Mathematical Problem Solving With the MATH Dataset}, 
      author={Dan Hendrycks and Collin Burns and Saurav Kadavath and Akul Arora and Steven Basart and Eric Tang and Dawn Song and Jacob Steinhardt},
      year={2021},
      eprint={2103.03874},
      archivePrefix={arXiv},
      primaryClass={cs.LG},
      url={https://arxiv.org/abs/2103.03874}, 
}

@article{threadweaver2025,
  title={ThreadWeaver: Adaptive Threading for Efficient Parallel Reasoning in Language Models},
  author={Lian, Long and Wang, Sida and Juefei-Xu, Felix and Fu, Tsu-Jui and Li, Xiuyu and Yala, Adam and Darrell, Trevor and Suhr, Alane and Tian, Yuandong and Lin, Xi Victoria},
  journal={arXiv preprint arXiv:2501.00000},
  year={2025}
}

@article{besta2023graph,
  title={Graph of thoughts: Solving elaborate problems with large language models},
  author={Besta, Maciej and Blach, Nils and Kubicek, Ales and Gerstenberger, Robert and Gianinazzi, Lukas and Gajda, Joanna and Lehmann, Tomasz and Podstawski, Michal and Nuelle, Hubert and Hoffra, Niklas and others},
  journal={arXiv preprint arXiv:2308.09687},
  year={2023}
}

@misc{alphaproof2024google,
  title={AI solves IMO problems at silver medal level},
  author={Google DeepMind},
  year={2024},
  howpublished={\url{https://deepmind.google/blog/ai-solves-imo-problems-at-silver-medal-level/}},
  note={Accessed: 2024-07-25}
}

@misc{minimax2025m2,
  title        = {MiniMax M2: Open Weight Large Language Model},
  author       = {{MiniMax AI}},
  year         = {2025},
  howpublished = {\url{https://www.minimax.io/news/minimax-m2}}
}

@misc{moonshot2025kimi,
  title        = {Kimi K2: Open Agentic Intelligence},
  author       = {{Moonshot AI}},
  year         = {2025},
  howpublished = {\url{https://moonshotai.github.io/Kimi-K2/}}
}

@misc{anthropic2025claudeopus45,
  title        = {Introducing Claude Opus 4.5},
  author       = {{Anthropic}},
  year         = {2025},
  howpublished = {\url{https://www.anthropic.com/news/claude-opus-4-5}}
}

@misc{google2025gemini3pro,
  title        = {Introducing Gemini 3 Pro},
  author       = {{Google DeepMind}},
  year         = {2025},
  howpublished = {\url{https://deepmind.google/technologies/gemini/}}
}

@misc{openai2026gpt55,
  title        = {GPT-5.5},
  author       = {{OpenAI}},
  year         = {2026},
  howpublished = {\url{https://openai.com/}}
}

@misc{anthropic2025claudecode,
  title        = {Claude Code: Deep Coding at Terminal Velocity},
  author       = {{Anthropic}},
  year         = {2025},
  howpublished = {\url{https://www.anthropic.com/claude-code}}
}

@misc{openai2025codex,
  title        = {Codex: AI Coding Partner from OpenAI},
  author       = {{OpenAI}},
  year         = {2025},
  howpublished = {\url{https://openai.com/codex}}
}

@software{vonwerra2020trl,
  title   = {{TRL: Transformers Reinforcement Learning}},
  author  = {von Werra, Leandro and Belkada, Younes and Tunstall, Lewis and Beeching, Edward and Thrush, Tristan and Lambert, Nathan and Huang, Shengyi and Rasul, Kashif and Gallou\'{e}dec, Quentin},
  license = {Apache-2.0},
  url     = {https://github.com/huggingface/trl},
  year    = {2020}
}

\appendix
\newpage
\section{Limitations and Future Work}
\label{app:limitations}
Parason's training and evaluation mainly focus on mathematical reasoning. It remains unclear how well the same taxonomy, data curation pipeline, and PA-GRPO objective transfer to other domains, such as real-world agents. Second, our current experiments focus on 8B-scale models. This setting provides a controlled testbed for studying parallel reasoning, but it does not fully show how Parason behaves across model families and sizes. In future work, we will scale Parason to more backbones and larger models to study whether the same parallelism patterns and latency gains hold at broader scales.

\section{Prompts Used for Data Curation}
\label{app:prompt}

We use the following prompts to identify Trial and Subtask steps in LLM reasoning trajectories.

For open-source models, we first split each reasoning trajectory by \texttt{\textbackslash n\textbackslash n}. The first prompt reconstructs logical steps by deciding whether each segment starts a new subproblem or continues a previous one. The second prompt then classifies each logical step as either a Trial step or a Subtask step.

For commercial models, the released reasoning summaries already mark steps with ``***Step i***''. We therefore apply the second prompt directly to classify each marked step as Trial or Subtask.
\begin{tcolorbox}[
    enhanced,
    colback=white,
    colbacktitle=black,
    title={Reconstruction by Logical Steps},
    fontupper=\small,
    attach boxed title to top left={yshift=-2mm, xshift=2mm},
    boxed title style={sharp corners, boxrule=0pt},
    arc=2pt,
    drop shadow={black!50!white},
    left=8pt, right=8pt, top=8pt, bottom=8pt
]
\textbf{SYSTEM:}
\begin{Verbatim}[breaklines=true, breakanywhere=true]
You are a helpful assistant that analyzes chain-of-thought traces from reasoning models.

The chain of thought has been split into multiple steps. For each step, decide whether it starts a new subproblem or continues the previous one.

Use the content of each step to make the decision. If a step introduces a new concept, question, or task that is distinct from previous steps, it likely starts a new subproblem. If it builds on previous steps by adding details, explanations, reflections, or calculations about the same concept, it likely continues the previous subproblem.

More specifically, steps that start with "Alternatively", "Wait", or "But" are likely to start a new subproblem, while steps that start with "Therefore", "Thus", or "Consequently" are likely to continue the previous subproblem.
Your output should be in the following format:

Step i: [New Subproblem|Continue Previous Subproblem]

If a step continues a previous subproblem, report where that subproblem starts. For example, if Step 5 continues the subproblem that started at Step 3, output:
Step 5: Continue Previous Subproblem (started at Step 3)

IMPORTANT: Output the analysis in the specified format. Use one line per step. Do not include additional explanations or text.

Example:
Step 1: New Subproblem
Step 2: Continue Previous Subproblem (started at Step 1)
Step 3: New Subproblem
Step 4: Continue Previous Subproblem (started at Step 3)
Step 5: Continue Previous Subproblem (started at Step 3)
\end{Verbatim}
\end{tcolorbox}

\begin{tcolorbox}[
    enhanced,
    colback=white,
    colbacktitle=black,
    title={Parallel Stage Identification},
    fontupper=\small,
    attach boxed title to top left={yshift=-2mm, xshift=2mm},
    boxed title style={sharp corners, boxrule=0pt},
    arc=2pt,
    drop shadow={black!50!white},
    left=8pt, right=8pt, top=8pt, bottom=8pt
]
\textbf{SYSTEM:}
\begin{Verbatim}[breaklines=true, breakanywhere=true]
You are given a series of mathematical reasoning steps. Classify each step into one of the following categories:
1. Trial Step: A step that introduces a new idea, approach, or line of thought whose success is uncertain but worth exploring.
2. Subtask Step: A step that belongs to a known solution path and directly contributes to the final solution.

Your output should be in the following format:
Step i: [Trial Step|Subtask Step] [Reason]

The number of output steps should match the number of input steps. Each step has its own number and is separated by "================================================================================". Ignore the content under "Content after </think> tag:
"

Give the reason for each classification after the step type.
\end{Verbatim}
\end{tcolorbox}

\section{AIME 2024 Problem Indices}
\label{app:aime24_indices}

Table~\ref{tab:aime24_difficulty} lists the AIME 2024 validation problems used for the difficulty-based analysis in Table~\ref{tab:token_breakdown_parallelism}. We group problems into Easy, Mid, and Hard subsets according to their difficulty labels, and report the corresponding zero-based problem indices for reproducibility.

\begin{table}[h]
\centering
\begin{tabular}{ll}
\toprule
\textbf{Difficulty} & \textbf{Problem Indices} \\
\midrule
Hard & 2, 3, 13, 21, 28, 29 \\
Mid & 1, 4, 5, 10, 16, 17, 18, 20, 25, 26, 27 \\
Easy & 0, 6, 7, 8, 9, 11, 12, 14, 15, 19, 22, 23, 24 \\
\bottomrule
\end{tabular}
\caption{AIME 2024 Problem Classification by Difficulty.}
\label{tab:aime24_difficulty}
\end{table}


\end{document}